\documentclass{ieeeaccess}
\usepackage{cite}
\usepackage{amsmath,amssymb,amsfonts}
\usepackage{algorithmic}
\usepackage{graphicx}
\usepackage{textcomp}

\usepackage{hyperref}

\usepackage{listings}
\usepackage{color}
\definecolor{codegreen}{rgb}{0,0.6,0}
\definecolor{codegray}{rgb}{0.5,0.5,0.5}
\definecolor{codepurple}{rgb}{0.58,0,0.82}
\definecolor{backcolour}{rgb}{0.95,0.95,0.92}
\lstdefinestyle{mystyle}{
    backgroundcolor=\color{backcolour},   
    commentstyle=\color{codegreen},
    keywordstyle=\color{magenta},
    numberstyle=\tiny\color{codegray},
    stringstyle=\color{codepurple},
    basicstyle=\ttfamily\footnotesize,
    breakatwhitespace=false,         
    breaklines=true,                 
    captionpos=b,                    
    keepspaces=true,                 
    numbers=none,        
    numbersep=5pt,                  
    showspaces=false,                
    showstringspaces=false,
    showtabs=false,                  
    tabsize=2
}

\ifCLASSOPTIONcompsoc
\usepackage[caption=false, font=normalsize, labelfont=sf, textfont=sf]{subfig}
\else
\usepackage[caption=false, font=footnotesize]{subfig}

\def\BibTeX{{\rm B\kern-.05em{\sc i\kern-.025em b}\kern-.08em
    T\kern-.1667em\lower.7ex\hbox{E}\kern-.125emX}}
\begin{document}
\history{Date of publication xxxx 00, 0000, date of current version xxxx 00, 0000.}
\doi{10.1109/ACCESS.2017.DOI}

\title{KIT MOMA: A Mobile Machines Dataset}
\author{\uppercase{Yusheng Xiang}\authorrefmark{1,2,4}, \IEEEmembership{Student Member, IEEE},
\uppercase{ Hongzhe Wang\authorrefmark{1}, } 
\uppercase{ Tianqing Su}\authorrefmark{3},\IEEEmembership{Member, IEEE},
\uppercase{ Ruoyu Li}\authorrefmark{1},
\uppercase{ Christine Brach}\authorrefmark{2},\IEEEmembership{Member, IEEE},
\uppercase{ Samuel S. Mao}\authorrefmark{4},
\uppercase{ Marcus Geimer}\authorrefmark{1},\IEEEmembership{Member, IEEE},
}
\address[1]{Institute of Vehicle System Technology, Karlsruhe Institute of Technology, Karlsruhe, 76131 Germany (e-mail: marcus.geimer@kit.edu)}
\address[2]{Division of Mobile Hydraulics, Robert Bosch GmbH, Elchingen, 
89275  Germany (e-mail: christine.brach@boschrexroth.de)}
\address[3]{Institute of Communication Technology, Technical University of Braunschweig, 38106 Braunschweig, Germany 
(e-mail:  t.su@tubs.de)}
\address[4]{Department of Mechanical Engineering, University of California at Berkeley, Berkeley, CA, 94720, USA (e-mail: ssmao@berkeley.edu) }

\markboth
{Author \headeretal: Preparation of Papers for IEEE TRANSACTIONS and JOURNALS}
{Author \headeretal: Preparation of Papers for IEEE TRANSACTIONS and JOURNALS}

\corresp{Corresponding author: Yusheng Xiang (e-mail: yusheng.xiang@partner.kit.edu).}

\begin{abstract}

Mobile machines typically working in a closed site, have a high potential to utilize autonomous driving technology. However, vigorously thriving development and innovation are happening mostly in the area of passenger cars. In contrast, although there are also many research pieces about autonomous driving or working in mobile machines, a consensus about the SOTA solution is still not achieved. We believe that the most urgent problem that should be solved is the absence of a public and challenging visual dataset, which makes the results from different researches comparable. To address the problem, we publish the KIT MOMA dataset, including eight classes of commonly used mobile machines, which can be used as a benchmark to evaluate the SOTA algorithms to detect mobile construction machines. The view of the gathered images is outside of the mobile machines since we believe fixed cameras on the ground are more suitable if all the interesting machines are working in a closed site. Most of the images in KIT MOMA are in a real scene, whereas some of the images are from the official website of top construction machine companies. Also, we have evaluated the performance of YOLO v3 on our dataset, indicating that the SOTA computer vision algorithms already show an excellent performance for detecting the mobile machines in a specific working site. Together with the dataset, we also upload the trained weights, which can be directly used by engineers from the construction machine industry. The dataset, trained weights, and updates can be found on our Github. Moreover, the demo can be found on our \href{https://www.youtube.com/playlist?list=PLEONnECiIX9-KaqDB8RFCd9scw5lQkUy_}{Youtube}.

\end{abstract}

\begin{keywords}
Dataset, autonomous driving, mobile machines, field robotics, computer vision, benchmarks, object detection, tracking, KIT MOMA, construction machines

\end{keywords}

\titlepgskip=-15pt

\maketitle

\section{Introduction}
\label{sec:introduction}
\PARstart{T}{he} research on the fully and semi-automated driving mobile machines are prosperous in the past decades. Mostly, the introduction of novel technologies aims to increase productivity, enhance the safety of the workers, and reduce the cost of operation. Among these new contributions, computer vision has attracted the most significant attention. Thanks to the boom of the deep learning, recognition capability of artificial intelligence outperform human-level recognition for many tasks. 

In the case of mobile machines, which usually work in a closed campus, making the autonomous driving of the mobile machines a level four task according to the standard from SAE \cite{SAEinternational.2016}. Currently, there are a lot of significant deep learning methods to visually detect the objects of interest, such as YOLO v3 \cite{Redmon.2018}, Faster-RCNN \cite{Ren.2015}, which achieved an appealing trade-off between speed and accuracy. 

Without a doubt, a series of researchers in the field of construction machines have been explored the possibility of using computer vision technologies and deep learning to recognize mobile machines. Although many of them have claimed that they propel the SOTA performance to a new and higher level, they did the research based on their unpublished dataset, which makes the results not comparable among each other and thus not plausible. Unfortunately, until today, no well-known database containing common devices for mobile machines, such as excavators, wheel loaders, bulldozers, and dumpers, is published with easy access and can be downloaded directly. 
As we know, the success of deep learning mainly benefits from three aspects: the generation of large-scale datasets, the development of robust models, and many computing resources available. The absence of the dataset limits the development of autonomous driving or working of the mobile machines.

To avoid the paucity of well-annotated images about mobile machines in current public datasets, we create a specific dataset for mobile construction machines: the KIT MOMA. Here images from varying viewpoints, poses, partial occlusions, and changing the depth of field were collected. A diversity of eight common categories across 5,663 images was organized in the standard PASCAL VOC dataset. 19,977 object instances were labeled for the research in the dataset. Based on our challenging dataset, a comparison among different algorithms become persuasive.  Also, we anticipate spurring the mobile machine detection to a higher level with a well-prepared dataset. Figure \ref{fig:dataset_kitmoma} illustrates the samples inside of our dataset.

\Figure[t!](topskip=0pt, botskip=0pt, midskip=0pt)[width=7in]{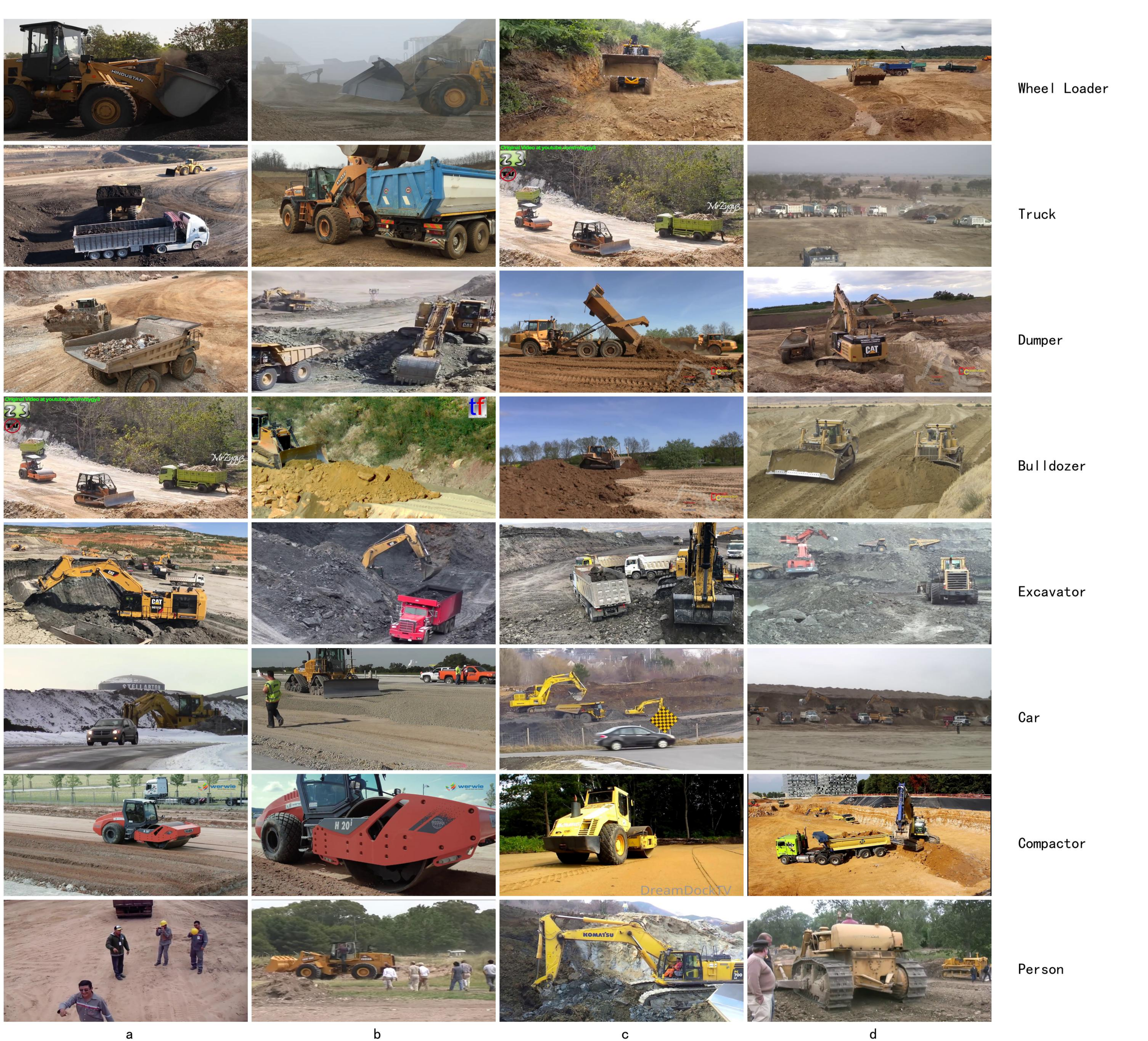}
{Sample images in dataset KIT MOMA, including 7 classes of construction machines as well as person with varying poses on the working scenarios. Images in column (a): objects in iconic view; column (b) objects in under partial occlusion; column (c) objects in varying poses; column (d) objects in non-iconic perspective.\label{fig:dataset_kitmoma}}

The contributions of this paper can be concluded as follow:
\begin{itemize}
    \item We publish a benchmark dataset suitable for detecting the commonly used mobile machines and comparing the performance of different algorithms to visual perception for mobile machines. The structure of our dataset is similar to PASCAL VOC for the convenience of the researchers in the fields of computer vision and deep learning.
    \item The KIT MOMA dataset is challenging: some instances are quite difficult to recognize and can only be detected with context information. There are many instances in figures since the working sites can be dense. We consciously selected these challenging figures to make the results tested on KIT MOMA plausible. The background of most of the figures in our dataset is the real construction site. In this fashion, we guarantee the same distribution between the training data and the test data to the greatest extent; thus, guarantee the performance of the winning model in the real-world application.
    \item Also, we publish the trained weights, which can be directly used to detect mobile machines to accelerate the implementation of computer vision technology in the construction machines industry. 
    \item We show that the current SOTA solution for common objects can also have a high performance to detect mobile machines with the help of our dataset, i.e., rather than considering the mobile machines as a specific task and try to develop more suitable algorithms, making an appropriate dataset can be an alternative to achieve the high-level detection task.
     \item As the tasks in the field of construction machines are level four, adding some custom figures of machines that need to be tested into our dataset can surely increase the predictor's performance. Thus, we also share the program to analyze the modified dataset. 
\end{itemize}

The rest of the paper is organized as follows: we first briefly introduce the previous studies of computer vision based algorithms and datasets for common objects and construction machines. We then present the KIT MOMA dataset in the following section with detail. Next, we analyze the performance of YOLOv3 on the KIT MOMA dataset and give the current feasible solution for the construction industry, i.e., show how to leverage our dataset and weights we offered to detect mobile machines in practical. Finally,  section  VIII gives conclusions and envisions the outlook.

\section{Related works}

\subsection{The well-known datasets}

Data are the prerequisite cornerstone of deep learning because deep learning models directly get knowledge from the data. Although the importance of the dataset is not so significant before 2012 \cite{Zhu.2012}, the deep-learning community has the consensus that the data have been the vital driving force behind computer vision technology \cite{Yu.2018, Milan.2016}. To circumvent the bottleneck of limited data, both Acuna and Yu proposed a method to accelerate the human labeling process by their software \cite{Acuna.2018} or a partially automated labeling scheme \cite{Yu.2015}. Besides these, Northcutt proposed an approach named Confident learning (CL) to evaluate the quality of the data \cite{Northcutt.2019}.  With the rapid development of computer vision, the dataset of image recognition is also enlarging at a rapid pace. For the classification, Caltech 256 is famous with more than 100 categories \cite{Griffin.2007}. Also, many scientists published a classification dataset based on videos, such as \cite{AbuElHaija.27.09.2016}. Besides that, more commonly applied datasets for general purpose are created, including PASCAL VOC dataset \cite{Everingham.2010}, Microsoft COCO \cite{Lin.2014}, and ImageNet \cite{Deng.20.06.200925.06.2009}. Also, there are a lot of specific datasets for specific tasks, including pedestrian \cite{Zhang.2017}, scene parsing \cite{Zhou.2017,Zhou.2014}, human activity \cite{CabaHeilbron.2015,Soomro.2012}, and face recognition \cite{Huang.2008}. 
In autonomous driving, KITTI is considered the pioneer, which contains objects of interest in the realistic scenarios of city Karlsruhe \cite{Geiger.2013}. Followed by Cityscape \cite{Cordts.2016}, RobotCar \cite{Maddern.2017} have also contributed to the autonomous driving community with their diverse dataset. Since the aforementioned automated driving datasets are collected in European countries, the scientists in the other parts of the world also published their dataset with much larger sizes, concretely they are BDD 100K \cite{Yu.2018} in the USA, ApolloScape \cite{Huang.2018} in China, and nuScenes \cite{Caesar.2019} in both the USA and Singapore.  

By the comprehensive literature review, we can conclude that a benchmark dataset should be diverse, abundant, consistent with the actual scene, and online release. 

\subsection{Recent object detection algorithms}

Computer vision is a grand and long-standing subject. Before 2012, the most outstanding algorithms are based on hand-crafted features, such as Histogram of Oriented Gradient (HOG) \cite{Dalal.2005} and Scale-invariant Feature Transform (SIFT) \cite{Lowe.1999}. In this period, a famous algorithm based on convolutional neural networks is LeNet \cite{Lecun.1998}. However, due to the limitation of the level of computer computing technology at the time, it is quite shallow and with too few training parameters. Therefore, the advantages of deep learning at that time is not significant. Nevertheless, after the AlexNet \cite{Krizhevsky.2012} won the ImageNet challenge \cite{Deng.20.06.200925.06.2009}, so-called large scale recognition challenge, in 2012, the deep convolutional neural networks have attracted much research attention in recent years. In 2014, the VGG-16 \cite{Simonyan.2014} was proposed, and it is used as the base network for many applications. After that, the inception network \cite{Szegedy.2016}, which combines the most of deep learning ideas, is designed. Among them, a particular form is called GoogleNet. Moreover, He shows that the neural networks can even surplus the human-level recognition \cite{He.2015}, and he invented the ResNet \cite{He.2016}, including the concept skip connection, making the training of much deeper neural networks possible, because the identity function is easy for the residual block to learn, in the same year. Usually, deep neural networks have a large number of training parameters and thus need plenty of time to be trained. To address this problem, transfer learning has got attention. The training time on the specific tasks can be dramatically reduced through transfer learning compared to if we train the whole model from scratch. Therefore, instead of directly training the total model, most of the researchers download the pre-trained ImageNet models. Until the time we write this paper, the most well known and successful computer vision algorithms to detect objects are YOLO, RCNN, SSD, and their variations. Redmon developed from YOLO \cite{Redmon.2016} to YOLO v2 \cite{Redmon.2017} and then to the YOLO v3 \cite{Redmon.2018}, whereas RCNN \cite{Girshick.2013} was enhanced to fast R-CNN \cite{Girshick.2015} and faster R-CNN \cite{Ren.2015}. The comparison among these algorithms was made and can be found in many scientific papers, such as \cite{Zhao.2019}: thus, here we only make a brief summary. Since YOLO v3 is a one-stage method and solves the task as a regression problem, it is quicker and famous for real-time capability. In contrast, faster RCNN adopts the region proposal network and achieve slightly higher accuracy in most competitions and tasks. In our paper, we evaluate our dataset based on the performance of YOLO v3 since YOLO v3 reduces the burden of hardware. 

\subsection{The previous contributions on detecting mobile machines}

To date, besides some common purpose, computer vision is used in many specific applications, such as airplane detection \cite{Chen.2018}, ship detection \cite{Wang.2018}, and of course, mobile construction machines. 

The idea of using a camera to recognize mobile machines visually is not novel. To the best of the authors' knowledge, the first research can be traced back to 1990 when Eldin want to use a camera to increase the productivity of construction of a state prison in the USA. Before the rise of very deep neural networks, a series of researchers have already reached some achievements in these fields. Azar has developed a model for non-rigid equipment of excavators detection and pose estimation in construction images and videos \cite{Azar.2012}. In 2011, Chi used a background subtraction algorithm to extract motion pixels, which are then grouped into regions. After that, the group will be identified using classifiers \cite{Chi.2011}. The dataset, comprising of 750 images, is equally divided into three classes: skid steer loader, backhoe, and worker. It achieved overall classification errors of 3.9\% with neural networks. The research also pointed out the similarities between loader and backhoe may cause worse performance. Both Park and Memarzadeh presented a method that can be concluded as a combination of HOG and the HSV color histogram, to localize construction workers or equipment in video frames \cite{Park.2012, Memarzadeh.2013}. In 2014, Tajeen mentioned in their paper that they built an image dataset for construction equipment recognition, including 300 images \cite{Tajeen.2014}. After the convolutional neural network (CNN) success, the application of the CNN-based object detection in detecting mobile machines and construction sites has been undertaking over the past decade. A consensus in the mobile construction machines industry has been built: for a variety of image recognition tasks, well-designed deep neural networks have far surpassed previous methods based on artificially designed image features. Fang uses Improved Faster Regions with Convolutional Neural Network Features (IFaster R-CNN) approach to detect the excavators and workers in realtime on their own dataset \cite{Fang.2018}. Kim did both the research about scene parsing \cite{Kim.2016} and objects detection \cite{Kim.2018} of construction machines. In their following researches, the estimated context information was used to reduce the cost of the earthmoving process \cite{Kim.2018b}. In 2019, Son used a very deep neural network to detect the workers in the working site, which was claimed to have yielded an accuracy of 91\% and 95\%, exceeding the SOTA descriptor in image target detection methods at that time. In his paper, he emphasizes the importance of varying poses and changing background \cite{SonH.ChoiH.SeongH.andKimC..2019}. Also, Son points out that the visibility of the equipment operator is inherently poor \cite{SonH.SeongH.ChoiH.andKimC..2019}, which is consistent with our point of view. Recently, Bang proposed an image augmentation method to enhance the performance of objects detector on construction sites, achieving a recall of 66.76\% and precision of 53.08\% experimentally on the UAV-based resources \cite{Bang.2019}.

Based on our literature review, we find that the research from Kim  \cite{Kim.2018}  who also aims to detect mobile machines is mostly similar to our research. Besides mobile machines detection, he claims that they have build up a dataset based on the images in the ImageNet.  Since the R-FCN is powerful and the dataset is relatively large, no wonder they can achieve excellent performance. However, the dataset can not be used as a benchmark for two reasons. First and foremost, the background of the samples gathered from ImageNet is mostly not a real working site. This makes the winning algorithms in this dataset may not have an excellent performance in practice due to the dramatic domain shift. Also, although the authors have mentioned that they can share the dataset, the dataset is not publicly available online, which is not in compliance with the baseline of the computer vision community. However, we encourage that they can also publish their dataset to complement our dataset.

\section{Why we created KIT MOMA dataset}

Unlike in the PASCAL VOC visual detection challenge, where a variety of algorithms tested on the same benchmark dataset, the results of current researchers on detecting mobile machines are conducted on their own dataset, resulting in a lack of contrasting acceptable persuasion. For this reason, we get the idea of setting up a benchmark dataset containing not only basic iconic data but also non-canonical images, providing a relatively reasonable benchmark of construction machines on the fly to validate the advancement of SOTA algorithms. Moreover, we realize that many machine learning breakthroughs occur only after a mature dataset was built. Thus, we built the mobile machines dataset to inspire and accelerate the new technologies in the field of construction machines.

\section{the KIT MOMA Dataset}

In this section, we first summarize our steps to build the dataset and then describe the details in subsections.
The dataset KIT MOMA is created as a specific dataset for commonly used mobile machines, which is challenging and diverse. There is one thing worth mentioning; we believe that the cameras in the construction site are more likely to be fixedly installed on the ground than on the driving construction machines. Because in most cases, construction machinery works within a limited range, making the configuration that install the cameras on the vehicle not more an inevitable method. In addition, the advantages of fixing the cameras on the ground are obvious. First and foremost, the cameras installed on the ground can provide the depth information from the figures with the appropriate calibration of the cameras. Here is the calibration process relatively easy since the coordinate among cameras is constant without vibration. Also, a wider angle of view and a cleaner lens can be achieved. The machines are usually surrounded by the dust during working resulting in the limitation of the vision.  Thus, we preferred to select the images gathered from a perspective outside of the mobile machines, which is quite different from the self-driving cars' training images. In this fashion, wireless communication should be developed for information sharing between machines and cameras. These researches can be found in \cite{Xiang.2020, Xiang.2020b}. Consequently, a diversity of eight common categories across 5663 images was organized in the form of the PASCAL VOC dataset. 19,977 object instances were labelled for the research in the paper, see Figure \ref{fig:instances_of_each_class}. 

\Figure[ht!](topskip=0pt, botskip=0pt, midskip=0pt)[width=3.3in]{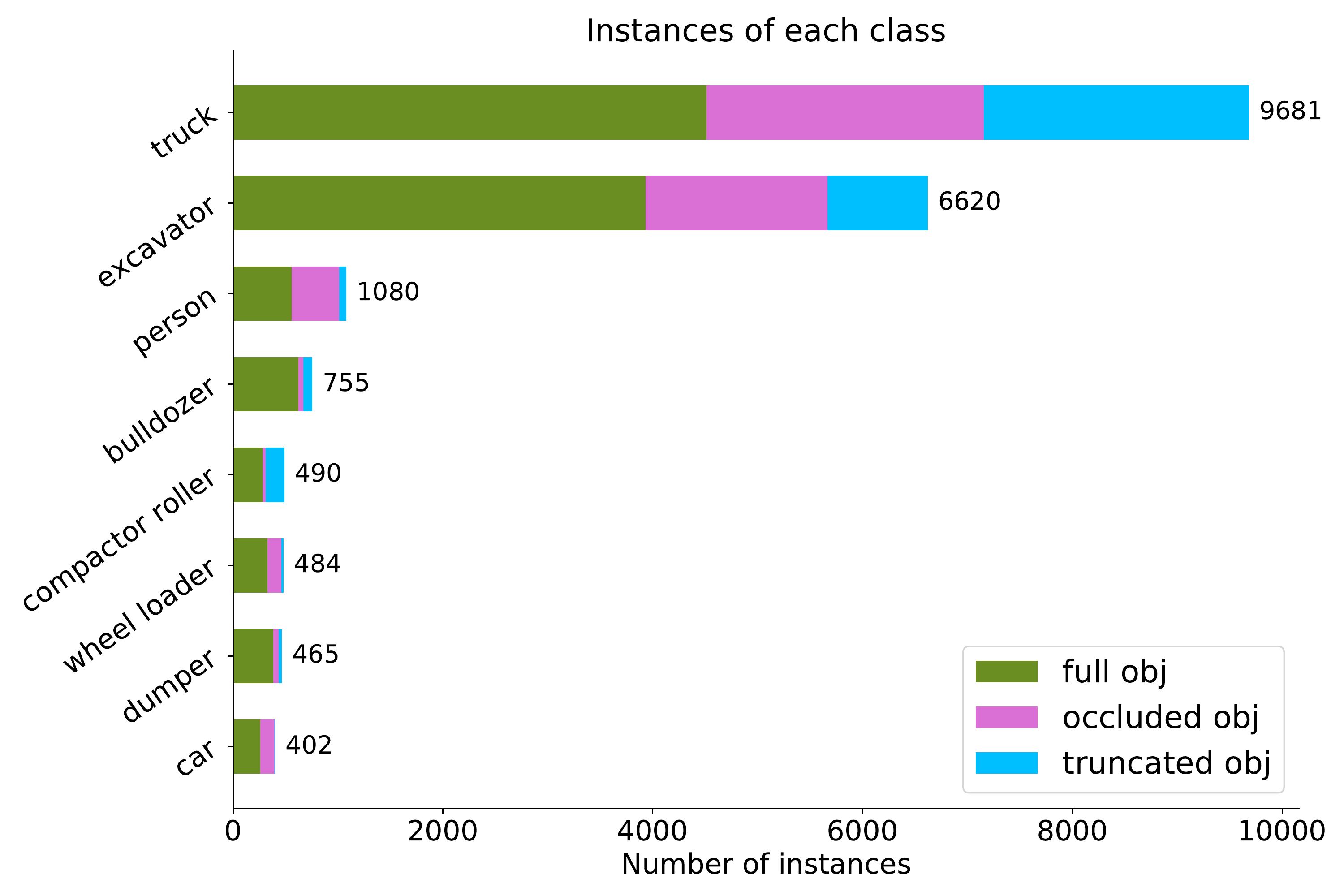}
{The statistics of KIT MOMA dataset. \label{fig:instances_of_each_class}}

Based on the survey about the most vital participants in the working site, we clearly defined the categories that we should focus on as the first step before collecting data. Unlike other categories, mobile machines vary to a certain extent depending on the components and working conditions.  On top of that, we human beings must be included in the dataset since most of the researchers believe the accurate detection of humans in the working site can improve security. Therefore we limited the species of detection tasks to common representative groups: excavator, truck, dumper, bulldozer, wheel loader, car, compactor roller, and person.

In order to guarantee the winning algorithms on our dataset can really have the best performance in practice, we collected the candidate images both from video frames and the official website of construction machines. The streaming video files are collected under the different real scenarios, which makes our dataset closer to the actual situation on the working site, and we then cut them into images. Besides the images from the videos, we also gathered some figures directly from the website of famous construction machines companies, such as Caterpillar, Komatsu, with the help of chromedriver and web crawler, since we believe that introducing these figures can enhance the performance of the predictors. Apparently, the figures from the videos are in a non-iconic view like the figures in MS COCO. In contrast, the figures from the official website of construction machines companies are canonical perspective as the samples in Caltech. Both of them make significant contributions to ensure a relatively high recall. Finally, for the visual perception task, more than 25,000 images were gathered, from which 5,663 representatives were selected.

Following, we annotated the selected images to label the ground truths of determined classes. We use the annotation tools "labelImg", which is mainly for object detection labeling work from Lin \cite{Lin.2015}. The software can generate both XML files for Faster-RCNN and text files for YOLOv3. Since the XML file contains more information than txt, we save the dataset in XML and then transfer the XML into txt. As we know,  labeling effort helps a dataset stand out in the training evaluation and detecting performance as well, whereas missing labels, false annotations, even widely unbalanced instance distribution, and too many clutters impair the effectiveness and robustness of a dataset. Therefore, before the dataset is fed into training models, it is worth analyzing the dataset by means of statistics and subsequently split it into subsets aiming to train the predictor and cross-validation. Whenever the dataset shows a significant imbalance among the interested categories, it would probably weaken the performance as a result. In this case, countermeasures such as label deficiency examination and then moderate supplement must be taken to keep the predictor robust against all classes. After careful preparation, KIT MOMA basically does not have such a problem; however, considering that some readers might add an additional class into our dataset, we also publish the code to evaluate the balance of classes in the dataset.  

Besides the balance among different classes, it is quite necessary to have the right balance between training and test set to gain a stable estimation of predictor performance. With less training data, the trained model tends to have a bias problem. In contrast, less testing data will lead to higher variance concerning the performance statistics. We randomly split the dataset into trainval (training and validation) and testing by a ration of 4:1. 

Finally, the richly-annotated dataset will be tested by the SOTA object detecting algorithm, concretely, YOLOv3. In the meantime, whenever we find that the detectors do not work well for a specific situation, we increase the number of labeled images in that case into our dataset. In this fashion, we increase the diversity and scene variation of the dataset. In addition, the metric mean Average Precision (mAP) is used to evaluate the detection performance. Here we use the recommended parameters and thus set the threshold of Intersection over Union (IoU) as 0.5. An AP (average precision) comparison with the best parameter settings will be conducted across all selected categories.

\subsection{Data Acquisition}

Thousands of images can be easily acquired as we have open access to a search engine and social media, e.g., Google and Flickr. Web images can be found and downloaded by crawling through websites. Hence, a scrapy crawler framework was built to grab pictures from Google search engine and mechanical engineering machinery websites. Special python scripts for each provider were created based on their site's HTML structure. By executing the python file we created, images of interests from all pages on the website can be collected.

Nevertheless, most search engine based images present a canonical view of objects, which could bias the algorithm to assume mobile machines are always located at the center view. This may lead to a deviation from the predictors' optimal performance if they are trained only with these images. Despite their weakness in the real inference, the web-based images from various providers show diversity in size, luminance, resolution, color, background, as well as ambiguity and thus help models gain an understanding of essential object features. Moreover, in fact, most construction machines providers publish their new models timely on their website; thus, adding these figures can enhance the predictors' recognition capability. Since these figures are quite easy to be detected and thus may exaggerate the performance of detectors, they do not include in the main dataset of KIT MOMA. However, they are well prepared and saved in the additional file in our dataset for use. 

By demonstrating the multi-angle and realtime working status of mobile machines, videos strengthen the generalization of predictors in realistic working surroundings. By appropriately extracting the images from videos every 50 fps records, we build up the non-iconic part of the dataset. In this part, we consciously select the videos varying the machinery working poses as well as the machine size due to the depth of the perspectives. Since the images are collected from realistic scenarios, occlusion and truncation are inevitable. In this fashion, thousands of images can be produced, making the detector feasible in various practical scenarios and, of course, realtime detection. A volume of 20895 images was captured from 125 videos, and 5663 from them were picked out for training the models and their validation. 


\subsection{Dataset format}

We propose PASCAL VOC format as the exemplar dataset format for our task. Figure \ref{fig:structure_kitmoma} illustrates the structure of KIT MOMA. 

\Figure[ht!](topskip=0pt, botskip=0pt, midskip=0pt)[width=3.3in]{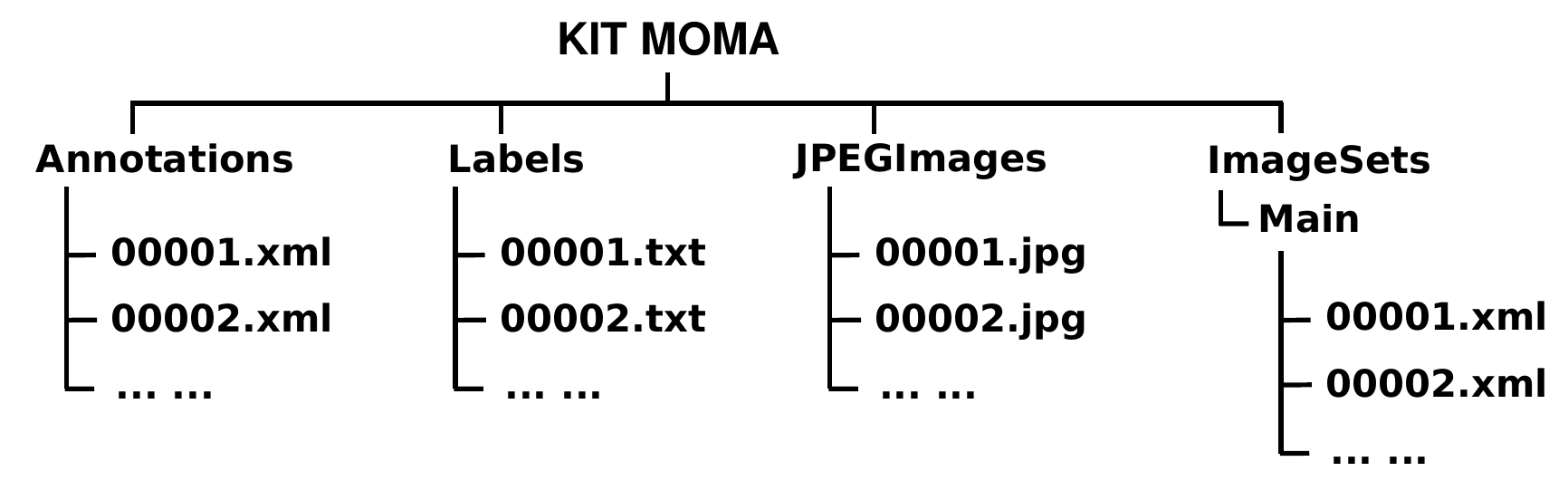}
{Hierarchical structure of KITMoMa, based on PASCAL VOC \label{fig:structure_kitmoma}}

Similar to PASCAL VOC, directories "Annotations", "labels", "JPEGImages", and subdirectory "Main" under "ImageSets" are the essential components, with relevant files in them. During the implementation of Faster-RCNN on KIT MOMA, file types such as XML, jpg, and files train.txt and test.txt in "Main" are in the necessity, while YOLOv3 detector will be trained with label text files, jpg files, and train.txt and test.txt in folder "Main".

All the links about KIT MOMA are well available in our GitHub repository. Labeling is non-trivial work, to avoid duplicating the creation of rectangular boxes and annotating them, we built the dataset only in the format of XML. SOTA object detection algorithms such as Faster-RCNN, SSD, YOLOv3, etc. require basically the same essential annotation information of targets of interests in two-dimensional images, including their coordinates and categories, which are generally expressed in the form of (left, top. width, height, and class). Although ground-truth targets were merely labeled in the format of XML to save labor work, text annotation can be transformed by program correspondingly, which is also available in our Github. During the transformation, the location information for every objects is translated from $(x_{min}, y_{min}, x_{max}, y_{max})$  to  $(x_{center}, y_{center}, w, h)$ to fit the two algorithms respectively. Besides, in the annotation, all the coordinates, width, and height are normalized, range from 0 to 1. Therefore attention should be paid whenever parameters for $x,y,w,h $  are calculated. For instance, a constant image size must be multiplied in the optimization process by k-means clustering of the annotated anchors, because the anchor centroids are measured in pixels.

\subsection{Manual Annotation}

Labeling is exhausting and costly to perform but is the prerequisite in the task of object detection; all the aforementioned annotation files such as XML files in the directory "Annotation" have been labeled manually. We use the label tool "labelImg", which is a famous graphical image annotation tool available in GitHub repository from Lin \cite{Lin.2015}, to accomplish the labeling job. We annotated every single object in an image with a bounding box, enclosing the ground truth of objects and marking the class each object belongs to. Figure \ref{fig:labelimg} illustrates the graphic interface of the label tool "labelImg" and an annotation sample.

\Figure[t!](topskip=0pt, botskip=0pt, midskip=0pt)[width=3.3in]{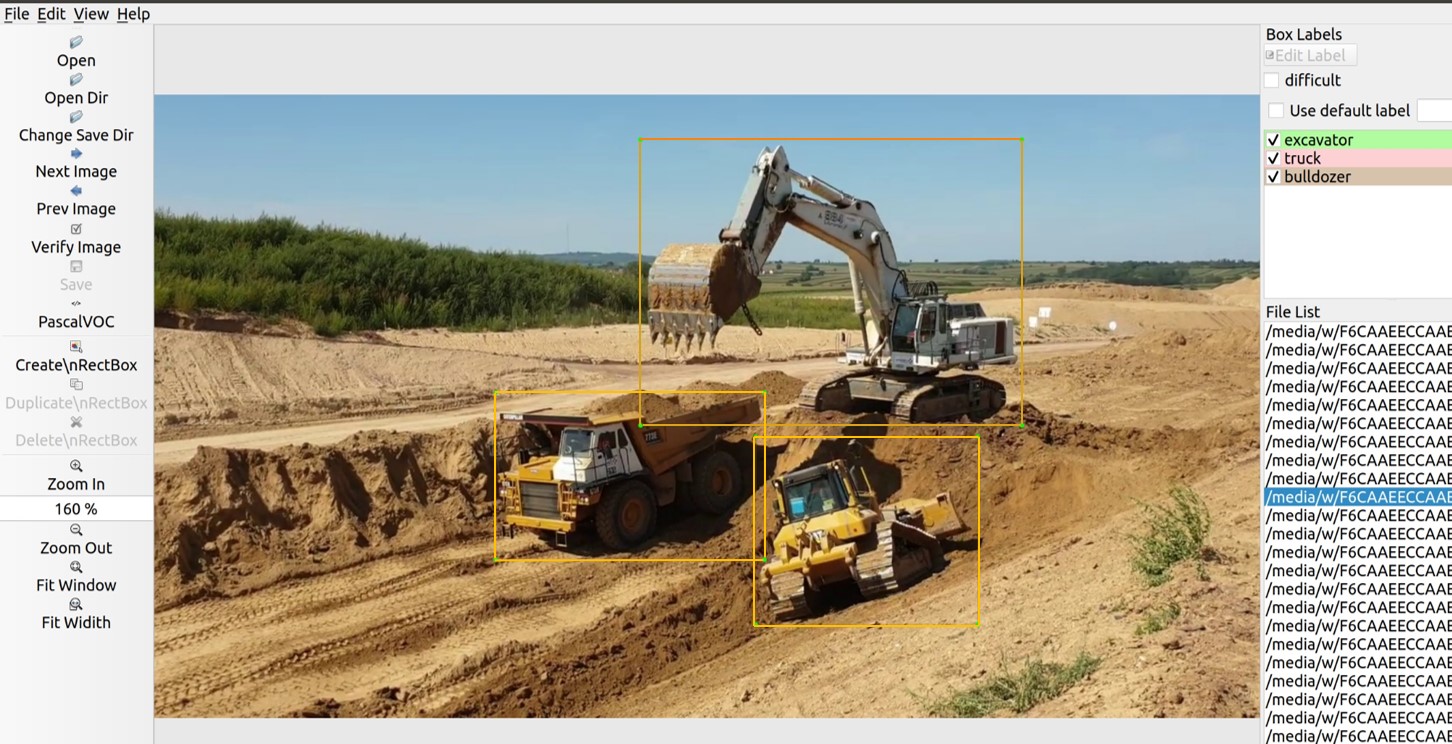}
{Label tool "labelImg" can load multiple images under a directory by clicking "Open Dir" on the menu, and save under pre-defined path. The shown bounding boxes that closely surround object ground truths, which the target instance is excavator in this figure, were made bold for the salience. Multi-class and multi-label for one single instance are possible; all marked labels are at the top right corners. As we have considered the PASCAL VOC format as the standard format for the KIT MOMA, all annotation files are saved in XML format \label{fig:labelimg}}

The saved XML file for the image annotated as in Fig. \ref{fig:labelimg} is represented in following code. It comprises all the ground truth information that we need to train the neural network with the samples.

\begin{lstlisting}[language=html]
<annotation>
	<folder>images</folder>
	<filename>sample.jpg</filename>
	<path>\path\to\the\sample.jpg</path>
	<source>
		<database>Unknown</database>
	</source>
	<size>
		<width>1280</width>
		<height>720</height>
		<depth>3</depth>
	</size>
	<segmented>0</segmented>
	<object>
		<name>excavator</name>
		<pose>Unspecified</pose>
		<truncated>0</truncated>
		<difficult>0</difficult>
		<bndbox>
			<xmin>561</xmin>
			<ymin>52</ymin>
			<xmax>1001</xmax>
			<ymax>382</ymax>
		</bndbox>
	</object>
	<object>
		<name>truck</name>
		<pose>Unspecified</pose>
		<truncated>0</truncated>
		<difficult>0</difficult>
		<bndbox>
			<xmin>394</xmin>
			<ymin>344</ymin>
			<xmax>704</xmax>
			<ymax>537</ymax>
		</bndbox>
	</object>
	<object>
		<name>bulldozer</name>
		<pose>Unspecified</pose>
		<truncated>0</truncated>
		<difficult>0</difficult>
		<bndbox>
			<xmin>694</xmin>
			<ymin>395</ymin>
			<xmax>950</xmax>
			<ymax>612</ymax>
		</bndbox>
	</object>
</annotation>
\end{lstlisting}

Here we summarize the most decisive info tags in XML that should be kept when transforming the format into txt files. 

\begin{itemize}
	\item
	Filename: name of the image file, in accordance with the text file under the path "KIT MOMA/Annotations/".
	
	\item
	Size: width, height, and depth of the image. Depth refers to the three image color channels: red, green, and blue. Images of this size will be fed into the convolutional neural network models for training or detection.
	
	\item
	 Object: including the class of the object and location. The "bndbox" stands for the bounding box, which is expressed in four coordinates (xmin, ymin, xmax, ymax). If multiple objects fall into an image, then all their specific names and corresponding positions will be recorded in the XML file. For instance, in the XML we showed, there are three objects: excavator, truck, and bulldozer.

    \item 
    Difficult: the objects in an image may sometimes be quite challenging to be detected only with the current image even for humans. For instance, if a truck is away from the camera, it will gradually become smaller and smaller so that it will become unclear in the end. To recognize the unclear spot, we must use the previous frame, i.e., the contextual information. In this case, We label such a sample as difficult.
	 
\end{itemize}

To decrease possible interference by potential ubiquitous noise, every single object of interest in an image, including occlusions and truncated instances, is labeled with care whenever human eyes can spot them. Cases of occluded and truncated objects are also counted as ground truths. Here we follow the idea from Yu \cite{Yu.2018} that the images should be specially pointed out if the cases are occluded and truncated objects. Concretely, we annotated a truncated excavator as "excavator, t", and an occluded excavator as "excavator, o", since LabelImg does not have the function to give this selection. The purpose of this method is to propel more robust algorithms. For the implementation of YOLO or Faster RCNN, we also provide the program to cancel these suffixes.  

To ensure the quality of our dataset, consistent rules were made for the labeling process as in the following items:

\begin{itemize}
    \item [1.] The label box size must be appropriate, i.e., the rectangular box should wrap the target closely. The rectangular box needs to contain information that distinguishes between different types of targets.
    \item[2.] Although a ground truth target may be blocked, it still needs to be marked as long as the human eye can identify the target. This improves the generality of the model. In the actual application scenario, there will be many obscured targets that the model, even so, should detect.
    \item[3.] Small size targets can not be missed if they are identifiable. SOTA detection algorithms are capable of multi-scale object recognition; thus, annotated tiny size ground truths boost the detection performance consequently. Figure \ref{fig:labelled_0924_3_3} depicts a labeling specimen of a far-off excavator behind two persons.  
    \item[4.] Targets that human observers can not recognize should be ignored. Otherwise, they will mislead the neural network. Only when humans can recognize them with the help of context information, we will label them and mark them as difficult.   
\end{itemize}

\Figure[ht!](topskip=0pt, botskip=0pt, midskip=0pt)[width=3.3in]{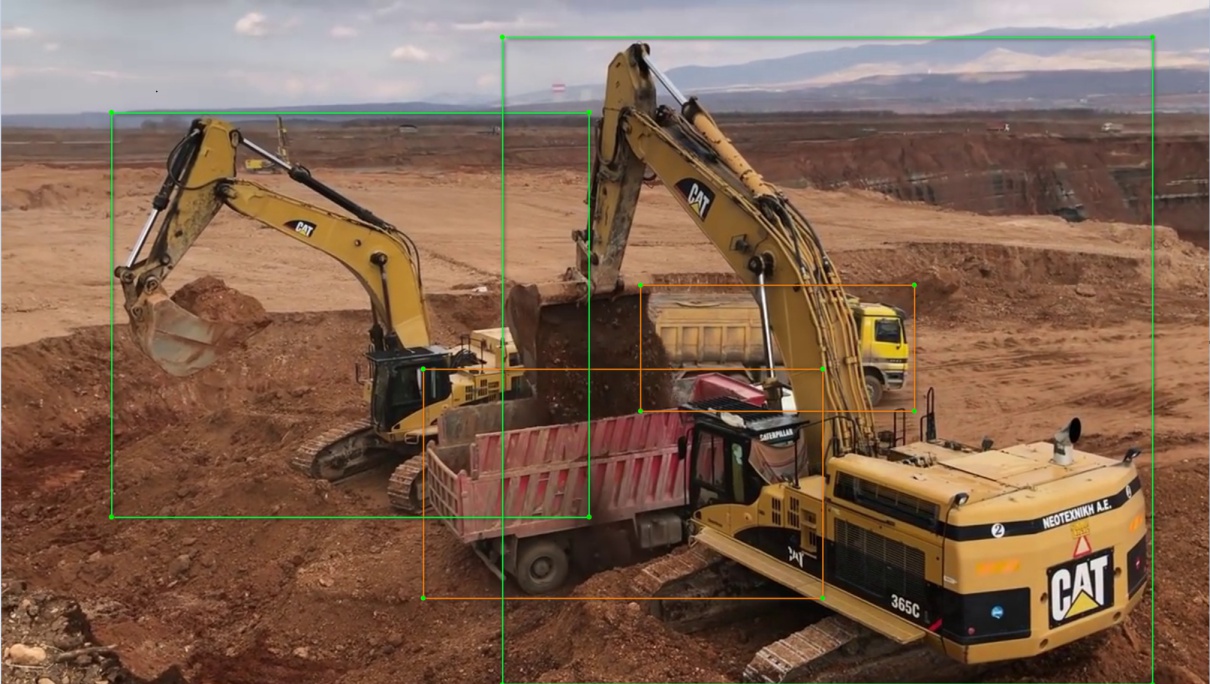}
{Two excavators and two trucks should be labeled in this image. They can all be distinguished by eyesight even though they are partially blocked. \label{fig:labelled-1-730}}

\Figure[ht!](topskip=0pt, botskip=0pt, midskip=0pt)[width=3.3in]{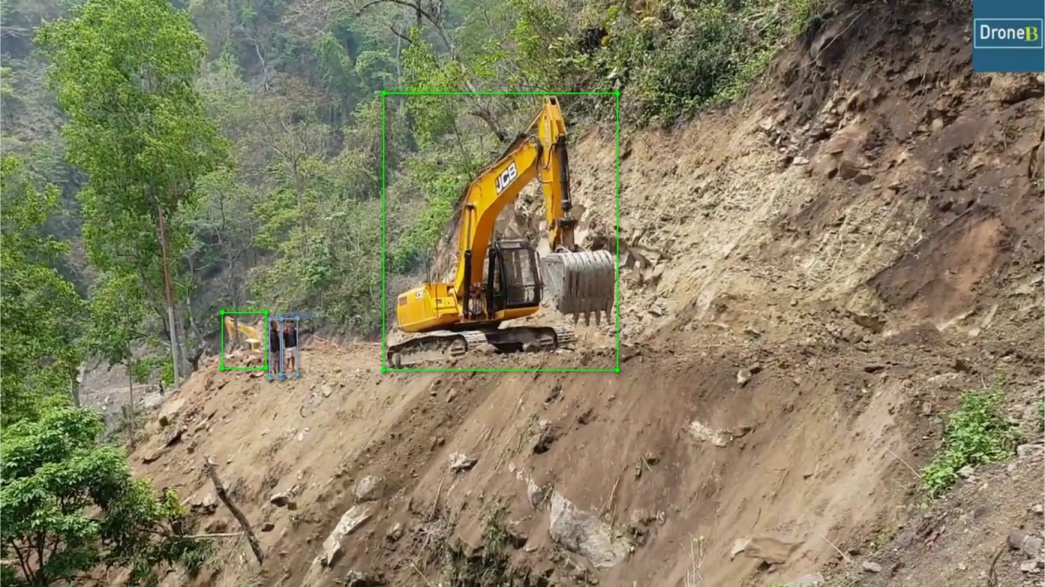}
{Four objects can be clearly seen in the image; even the excavator in the distance is much smaller than the one nearby. Moreover, the two standing workers can be recognized as well.\label{fig:labelled_0924_3_3}}

\subsection{Dataset splits}

As mentioned above, the dataset is meant for training as well as testing. Therefore we group it into four subsets randomly to ensure the training set and test set coincide in the data distribution. In order to achieve a stable estimation of model performance, a reasonable balance between training and test set is required. Depending on the volume of the database, it is quite flexible in determining the partitioning scale. Practically it gains better performance with a smaller proportion of testing set when the size of the dataset is larger. Based on our data amount, we split our dataset into approximately 80\% for training and validation, and 20\% for testing.

The arrangement of the data division is illustrated in Figure \ref{fig:structure_kitmoma}. In txt file like trainval.txt image file names with a suffix are stacked. Literately, all images in accordance with names in trainval.txt are intended for training and validation. Likewise, we test predictors using the images regarding the names in test.txt. Data inside trainval can be further split into training and validation subsets. Above all, these four groups work together to make full use of the complete database in order to gain satisfying predicting performance.

\subsection{Data Preprocess}

As a consensus, clean data helps improve detection performance. Prior to the implementation of CNNs, the dataset is analyzed statistically to strike out ineffective labels and ensure its conformity with the working scenarios. This is an essential step because we know that readers may modify our dataset to better suit their tasks.

To list annotated labels by the annotation tool "labelImg", a specific Python script was written. By executing the script, a list of class/count pairs would be printed, e.g. (excavator 536). In the case of typo error labels such as "excavater", further steps must be taken to rewrite the revised class into the XML files. The program which can automatically find the error is also on our Github. 

Moreover, difficult spotted instances should be averted for YOLO and faster RCNN. Figure \ref{fig:mis-label} depicts annotations of unrecognizable trucks and person, which mark as difficult and needs to be removed for YOLO and faster RCNN. In contrast, for other algorithms, these marks may be useful. 

\Figure[ht!](topskip=0pt, botskip=0pt, midskip=0pt)[width=3.3in]{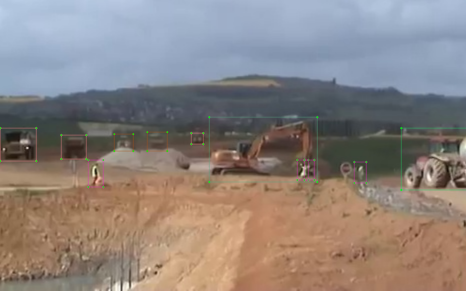}
{Rectangular correctly encompass ground truths: a dumper, an excavator as well as two persons. However, other trucks can be inferred from context stream frames but are not identifiable in the single image. The label should be ignored if we use YOLO v3 to detect mobile machines. \label{fig:mis-label}}

\Figure[t!](topskip=0pt, botskip=0pt, midskip=0pt)[width=7in]{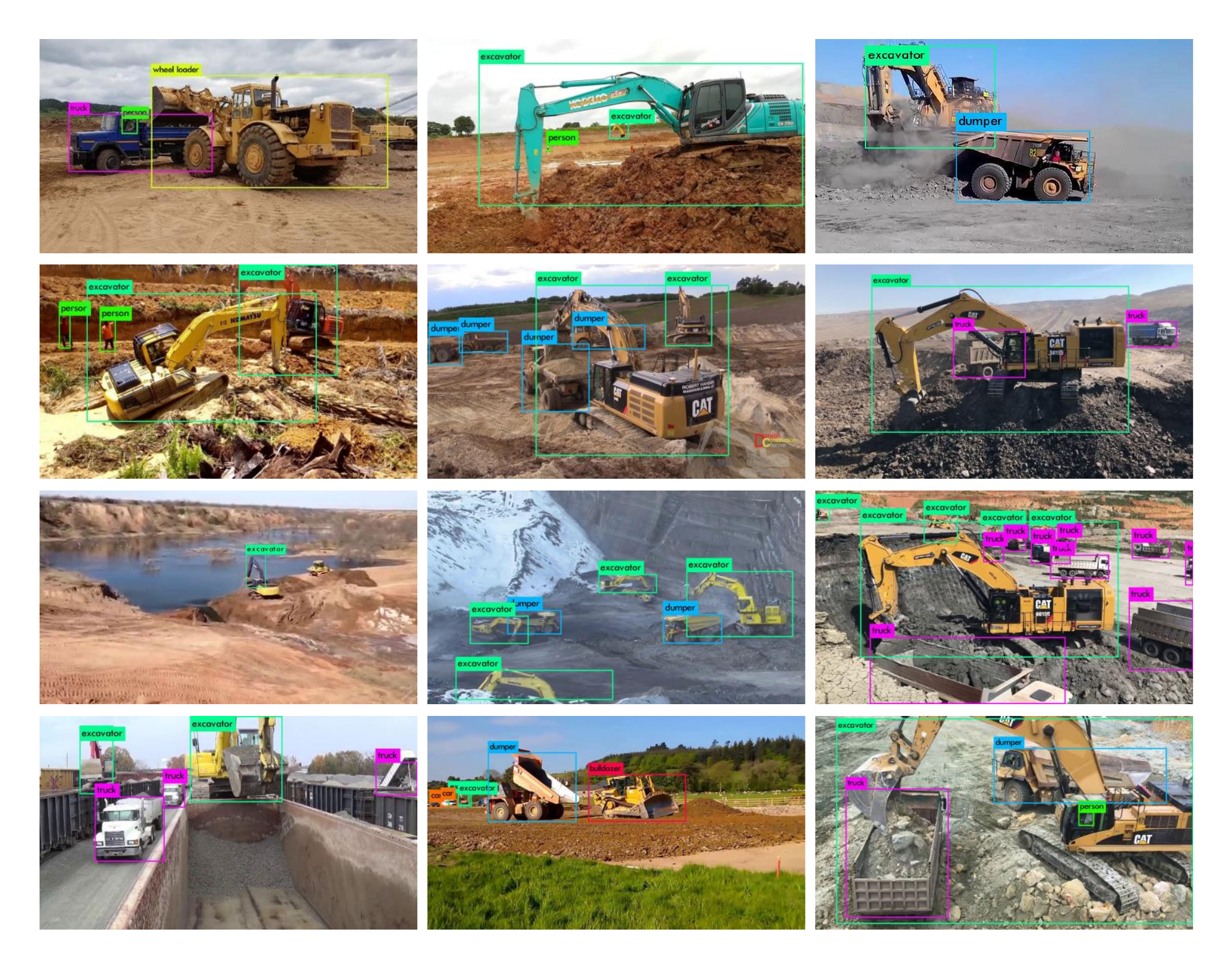}
{Prediction samples with optimized dataset and algorithm YOLOv3.\label{fig:Pred_MOMA_dataset}}

In this section, a special dataset KIT MOMA for the CNN-based visual perception of mobile machines was created, and preprocessing in necessity is also introduced to help the readers modify our dataset for their detection tasks. Instead of using our dataset to train the predictors from scratch, we recommend using Darknet-53 trained on ImageNet as a base network.

\section{Evaluation of the recent computer vision algorithm performance on KIT MOMA}

Although the initial goal of this dataset is for the scientists from computer vision to test and enhance their algorithms for detecting common mobile machines, we would like to encourage the engineers from construction machines to use the dataset and take advantage of the computer vision technologies for their application. Here, on the one hand, we evaluate our dataset by means of YOLO v3, and on the other hand, we show the model setup for the convenience of readers who just want to use our trained model directly. Since many mobile machines predictors have been built with Faster RCNN, we do not show the setup of Faster RCNN again to avoid redundancy. Here we only demonstrate the implementation of YOLO v3.

Object detection tasks demand high computational power, and for some practical cases such as video stream recognition, powerful computing devices are needed.  Since Google offers a graphic computing platform on which both models can be trained much faster than on a commonly used local laptop, we share the code on Google colab where the GPU is free to use. Nvidia GPUs boost the calculation by taking advantage of CUDA, a parallel computing platform that allows a graphics card to ameliorate a CPU's performance. Since mAP performance does not differ much, up to 1\%, between different GPU series even with different image scales or non-identical mini-batch sizes, it is ok to use our weights on other platforms. All the environments, including GPU, has been set up in our configuration files. We believe that even without deep learning knowledge and Linux can utilize the model. We trained to detect their own figures. The framework Faster-RCNN can reach a Frame Per Second (FPS) of 5, while YOLOv3 at about 45 on Tesla k80, which is offered for free by Google. As a comparison, a Nvidia GTX 1050 used in a mediocre laptop can achieve an FPS of 10 with YOLO v3.

The original YOLO algorithm was uploaded by Joseph Redmon on his website. Afterward, several revised versions came out in different programming languages and updated in quite different aspects. In our work, mostly the original version is applied. However, to extend some essential features, another prevailing repository is referred to as well, concretely, we use the version from Bochkovskiy, whose code can be found on his Github.

Ideally, for each category to detect, there should be at least one similar object in the training set, which should comprise likeness of shape, relative size, point of view, tilt, illumination, etc. of the targets. From that perspective, the larger the dataset, the better the detectors will be. However, it may take a couple of weeks to train the large dataset with the default settings in the configuration file. On this point, it might make the construction machine engineers flinch from the chance to use computer vision algorithms. Also, even with SOTA solutions, based on the test results on MS COCO with IoU of 0.5, the best mAP is about 50\%. Since our dataset is easier as MS COCO, the test results go to 85\%. However, it is surely unacceptable for the construction machines industry due to the safety reasons; it seems like those SOTA solutions should be improved for the detection of construction machines the same as the detection of cars. Alternatively, since the autonomous driving of construction machines is a level four task, which provides the possibility to increase the prediction performance by means of scarifying the generalization capability, i.e., the performance of the predictor for the specific working site with only limited kinds of mobile machines inside is more important than its performance to detect all the mobile machines in the world; thus, we focus on finding a current feasible solution for the construction machines industry in the following context. Generally speaking, if the distribution among the training, validation, and test dataset are the same, the predictor will perform its best performance. Besides, the mAP of the predictor can increase when we add some similar objects from different scenarios in the training data. Therefore, we recommend adding some additional annotated images of target mobile machines into our dataset and further train the model to get the optimal predictor for the level four task detection. To validate the idea, concretely, we take 666 well-annotated images from the KIT MOMA dataset into the network for training as well as validation. The basic idea of this approach is to increase the recognition rate of the target mobile machines by adding some samples of the target machines to be detected in a relatively small dataset to reduce the difficulty of detection. This approach is based on the assumption that no unexpected mobile machines will come to the working site. Obviously, for level four autonomous driving, this assumption is reasonable. The selected ground truth instances are plotted in the histogram in Figure \ref{fig:666-instances for 7 classes of interests}.

\Figure[ht!](topskip=0pt, botskip=0pt, midskip=0pt)[width=3.3in]{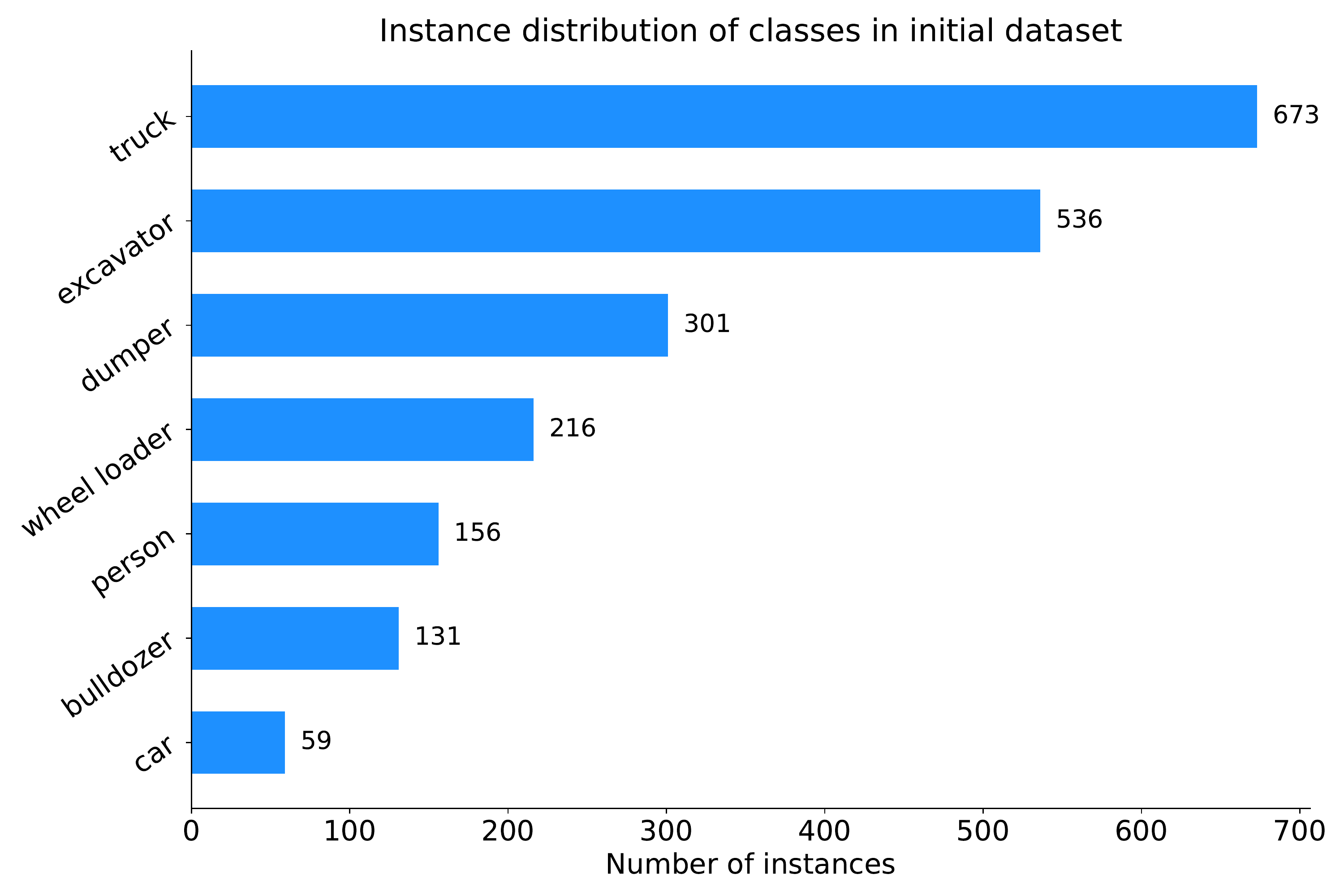}
{Class distribution on 666 images: the instances number of truck and excavator outnumber bulldozer and car since they attract more interest.\label{fig:666-instances for 7 classes of interests}} 

The training time is dramatically reduced, and the prediction results are illustrated in Figure \ref{fig:Pred_MOMA_dataset}. Notice that we uploaded both weight files trained on all figures in the dataset and trained on these 666 figures. The first one can make full use of the entire dataset to get better generalization ability, whereas the latter one is designed for the purpose that some engineers might add some custom figures and want to have a better performance on a special kind of object. 

On the images in Figure \ref{fig:pred_8000_iconics.jpg}, every single inference is marked with a bounding box in a different color to specify its category. Categories are labeled in the bounding box over the top left corner. The model appears to have satisfying performance on those images since they are in a canonical view and thus not so challenging.

\Figure[ht!](topskip=0pt, botskip=0pt, midskip=0pt)[width=3.3in]{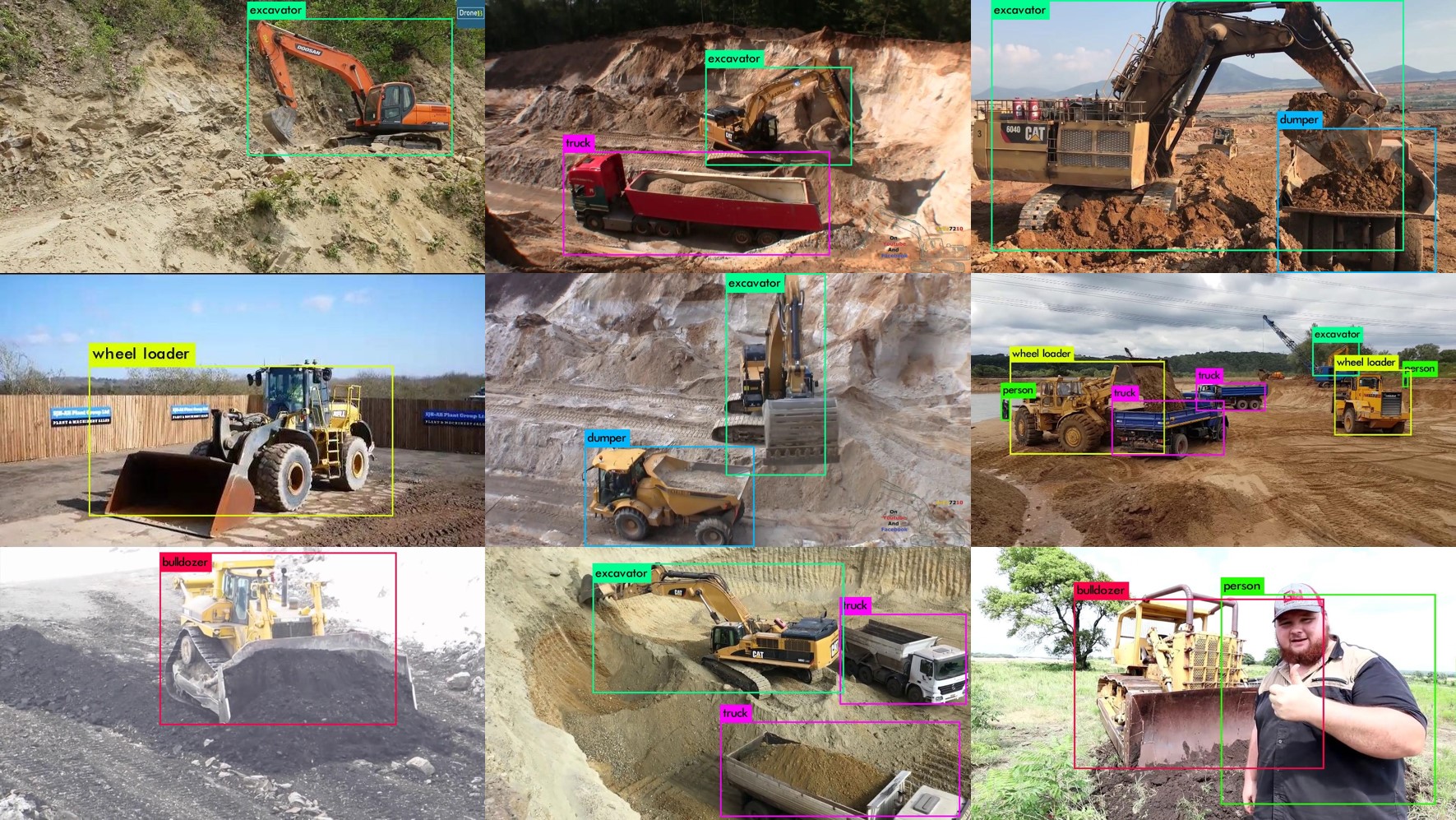}
{Sample of inference results by the 8000th predictor on images in iconic view. \label{fig:pred_8000_iconics.jpg}}

Although the predictor with default configuration can easily achieve excellent accuracy on iconic images, it can not have a satisfying performance on images in non-iconic view, which are not taken from a normal perspective, or with truncated, or blocked by other objects. An image can also become non-canonical when the whole image is obscured or ambiguous, or targets such as excavators are working surprisingly, e.g., sitting in the water. With the default setting of YOLO v3, the optimal performance may not be achieved. To address this problem, here we would like to share some useful tricks to improve the training process and the mAP of the YOLO v3 algorithm. 

First of all, by comparing the results from Figure \ref{fig:mAP_vs_batches_416_noroad} and Figure \ref{fig:mAP_vs_batches_416_withroad}, higher mAP performance can be achieved with a relatively balanced training dataset. 

\Figure[ht!](topskip=0pt, botskip=0pt, midskip=0pt)[width=3.3in]{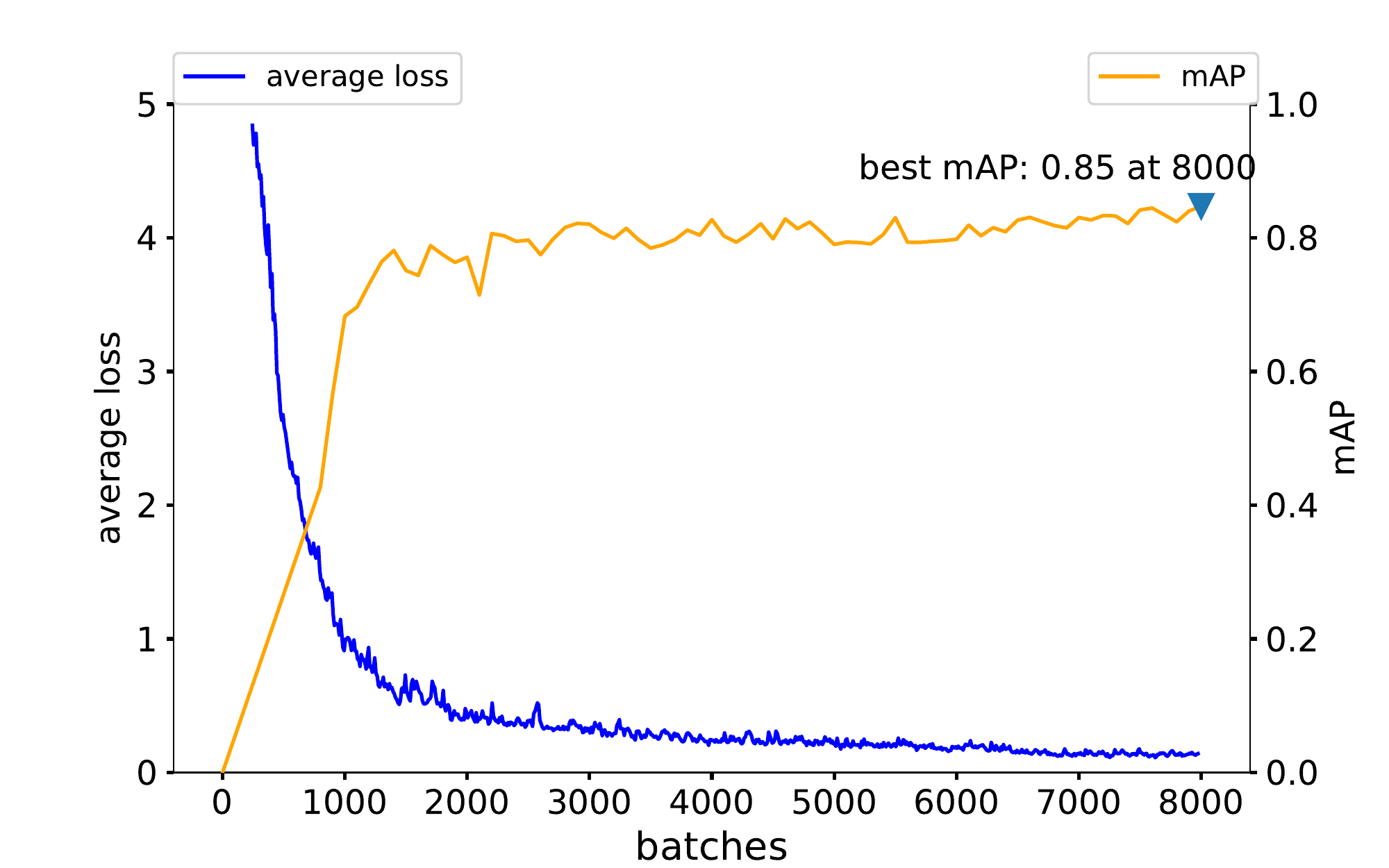}
{MAP over batches, trained with a balanced dataset.\label{fig:mAP_vs_batches_416_noroad}}

\Figure[ht!](topskip=0pt, botskip=0pt, midskip=0pt)[width=3.3in]{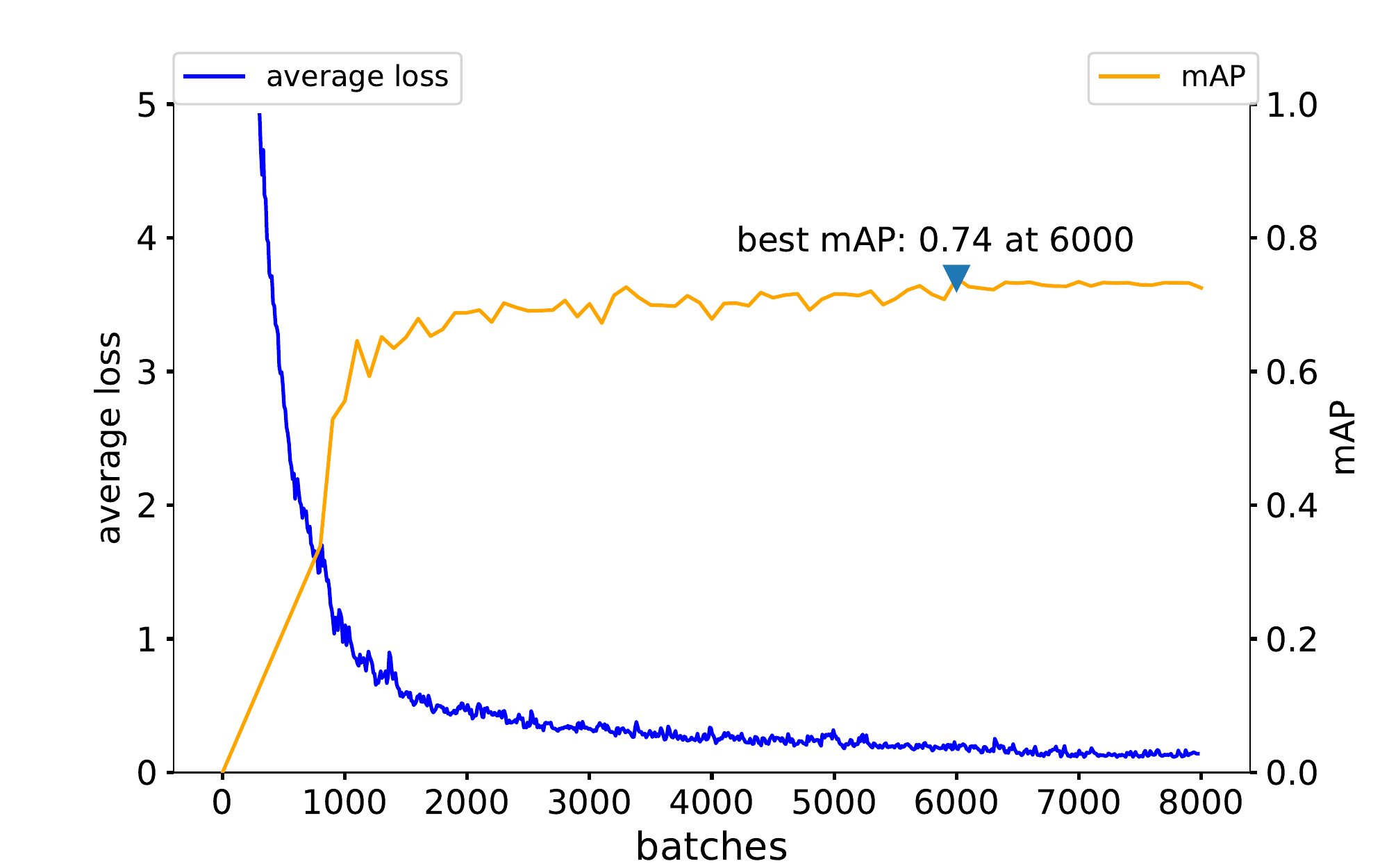}
{MAP over batches, trained with an unbalanced dataset. \label{fig:mAP_vs_batches_416_withroad}}

Second, according to the setting of YOLO, the multi-scale prediction is applied in feature maps. To narrow down the computing without hurting the prediction performance, we implement k-means to cluster the centroids of the positions of all the labeled objects. Instead of using the default anchor for the dataset COCO, we generate nine anchors as (16.0,26.0), (40.0,40.2), (30.8,84.4), (71.8,84.2), (119.6,124.2), (105.0,219.0), (191.6,175.2), (200.0,290.6), (322.6,346.6). 

\Figure[ht!](topskip=0pt, botskip=0pt, midskip=0pt)[width=3.3in]{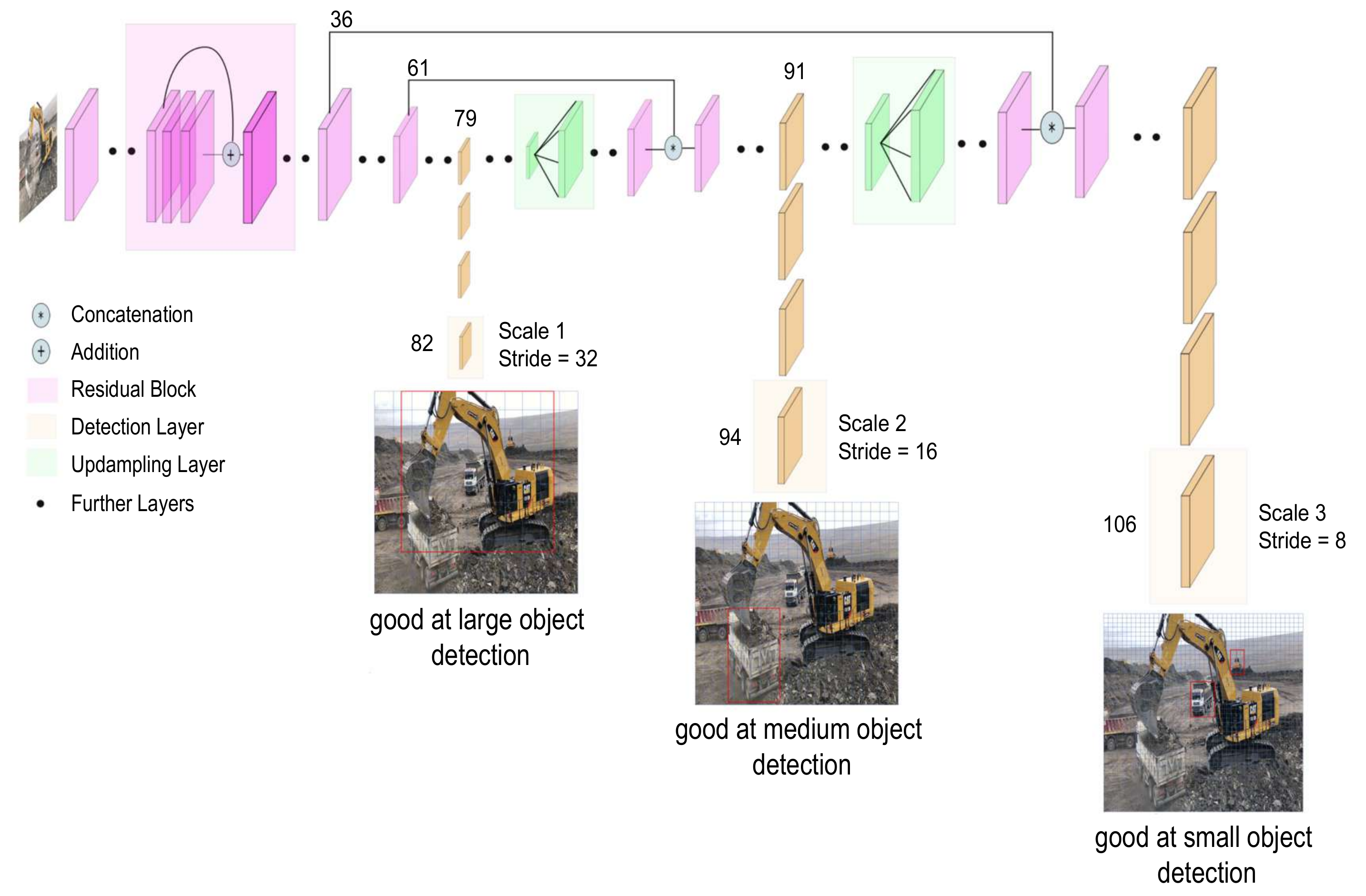}
{Hierarchy of predictor with skip connections, e.g. 94th layer, responsible for detecting medium-size targets, relates to the 61st layer before downsampling. Likewise, 36th layer is directly connected to 91st layer by a short cut.\label{fig:multi-scale-detection}}

Third, following the expectation that more training batches but with smaller learning rates could improve detection performance, we decay the training steps after the average loss begins to fluctuate. Concretely, we set the learning rate as follows,

\begin{lstlisting}
steps=8000,10500,12000
scales=.5,.1,.1
\end{lstlisting}
Here step learning rate decay of 0.5, 0.1, and 0.1 are applied at the 8000th, 10500th and 12000th step, respectively. Usually, it would be sufficient with 2,000 batches for each class, and no less than 4,000 iterations in total, training work can be then stopped. Also, the learning process can also be stopped when the average loss no longer decreases. After 12000 training steps, the average loss function converges to no more than 0.1,  a quite adequate condition to stop.

With the new predictor, we run inferences on the no-iconic images, which are shown in Figure \ref{fig:pred_12000_noniconic}. Mobile Machines like excavators are usually in large size, and predictors may quickly get used to that dimension; however, if target excavators are zoomed out or seen from an irregular perspective, they can be small objects as well. Based on our experiments, the inference ability concerning classification and localization of the predictor on the first three images has been greatly enhanced compared to the default setting.  It remains blind to the excavator in the last image of Figure \ref{fig:pred_12000_noniconic}. Regrettably, nothing is found even though human observers can easily discern the mobile machine (an excavator) on the left. Possible reasons for that are the lacking of remarkable characteristics and its tiny size. More images of this size and pose should be trained to improve the identification capability. Although some instances are still not detected, the holistic performance of the predictor is satisfying since a shorter range deserves more attention.

\Figure[ht!](topskip=0pt, botskip=0pt, midskip=0pt)[width=3.3in]{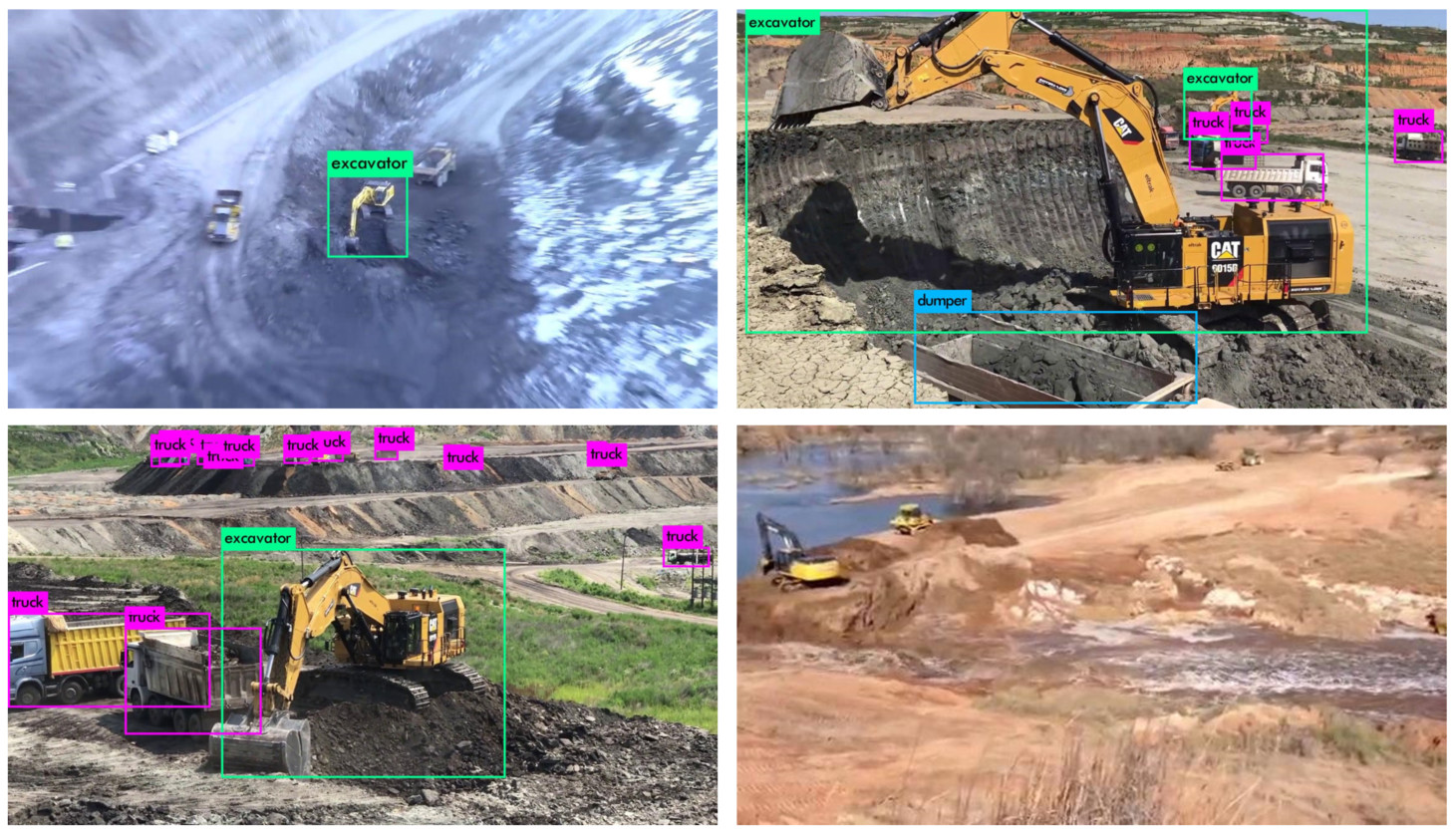}
{Contrast inference by predictor at the 12000th batch made on images with a non-iconic view.\label{fig:pred_12000_noniconic}}

Further differences between the three predictors at the 1900th, 8000th, and 12000th are shown in the $2 \times 4$ image grid. In Figure \ref{fig:contrast_best_8000_colab12000}, a and b are raw images, a1 and b1 are predicted by the first predictor at the 1900th batch. Similarly, a2 and b2 are from the second predictor at 8000th. At bottom a3 and b3 are from the third predictor at 12000th batch. Apparently, the bounding box surrounds "dumper" closer as the training steps increase in Figure \ref{fig:contrast_best_8000_colab12000}, 
indicating that the predictor has 12000 batches acquired a more powerful ability to localize the targets.  Besides that, the dumper, which is in the blue bounding box, is recognized, and the fictitious noise of the truck is eliminated, which implies that the class confidence increases with the more trained predictor.

\Figure[ht!](topskip=0pt, botskip=0pt, midskip=0pt)[width=3.3in]{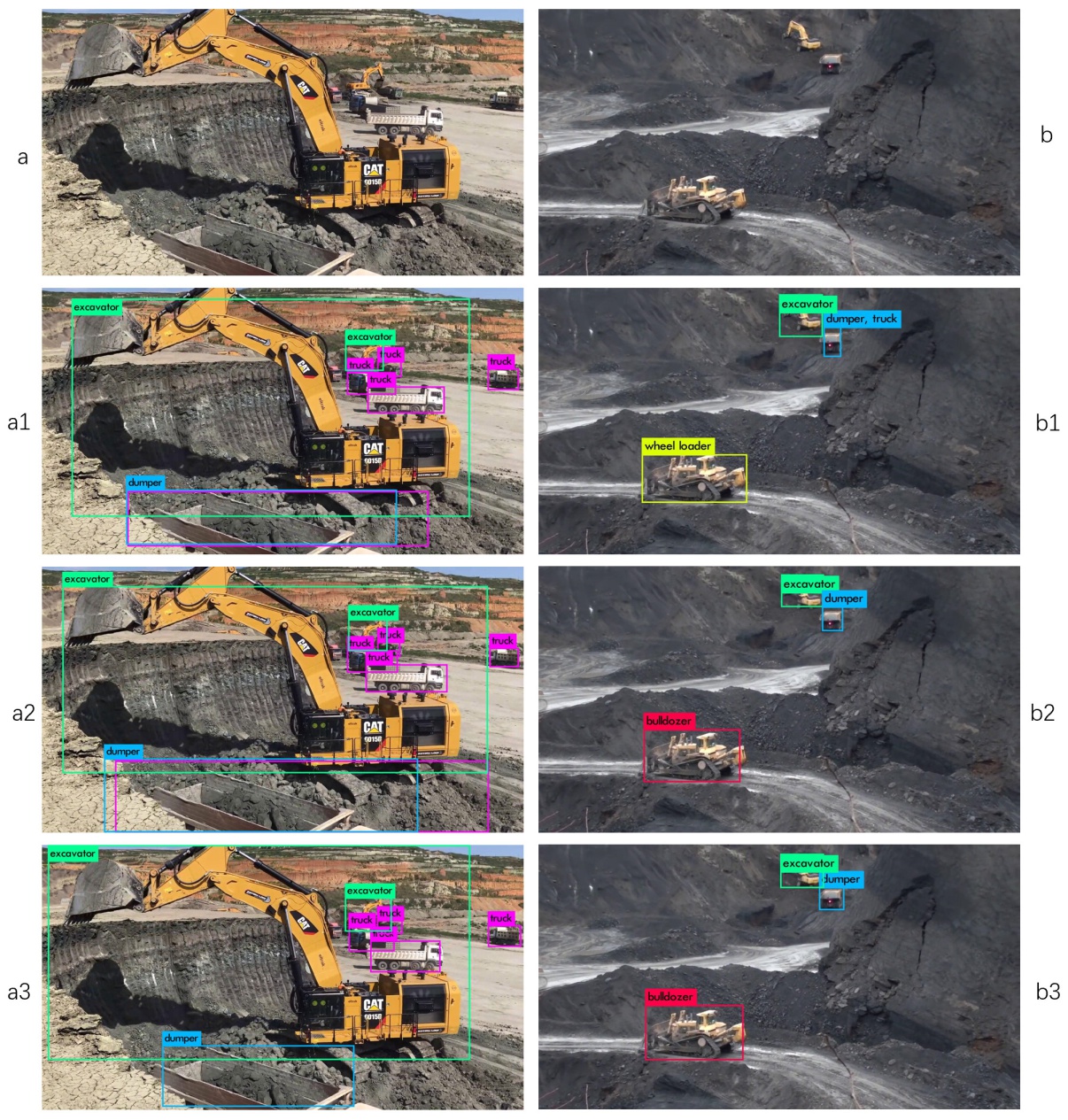}
{Prediction performance contrast by three predictors, which were made on images with both iconic view and non-iconic view. \label{fig:contrast_best_8000_colab12000}}

Figure \ref{fig:prediction on val by 1900, 8000, level4} shows the specific AP values on each class predicted by predictors under the different situations. Their trend illustrates AP increases with more batches for most of classes. Here is the test data quite similar to the validation data; hence, the predictors may overfit to the mobile machines that exist in the training and validation data. Although these results exaggerate the algorithm's real ability, it can accurately reflect its performance on the fourth level of autonomous driving.

\Figure[ht!](topskip=0pt, botskip=0pt, midskip=0pt)[width=3.3in]{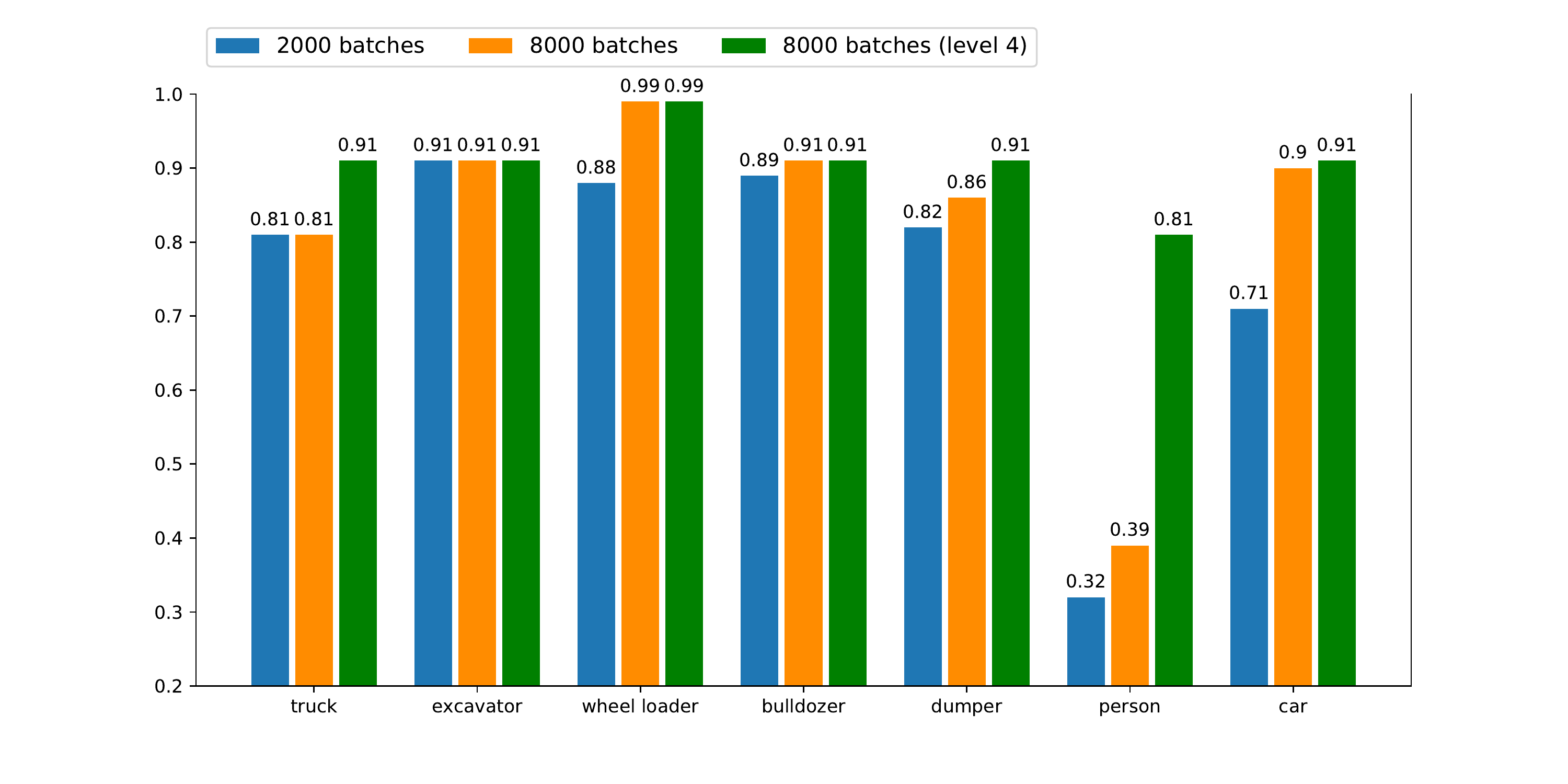}
{Individual prediction AP for each class made by predictors at batch 1900 in blue, 8000 in orange. The green Column demonstrates the performance if we take the assumption that only known mobile machines are working in the working site. i.e. level 4 autonomous driving case. \label{fig:prediction on val by 1900, 8000, level4}}

Although it might make no sense to show the generalization capability of the predictors since the assumption that no unexpected mobile machines are in the working site is reasonable, we further tested the predictor with 8000 batches on other 5,663 images in the KIT MOMA because we would like to show the method to solve the problem if AP is in some case unsatisfying low. From Figure \ref{fig:mAPfor5663}, we can see that the AP for each class goes much lower. Although the classes person and car are the lowest, it is predictable since we have fewer samples in these two classes. As a counter measurement, we can add samples from other datasets, and thus it cannot lead to a problem. The other colossal gap is the wheel loader. By analyzing the precision-recall curve, we found that the false-positive dramatically increases as the confidence threshold decreases, resulting in an extremely low AP. Moreover, we further analyzed the false detected samples. We found that most mistakes are the excavators with a shovel facing forwards since they are reconfigured for mines, or some trucks are very close to the cameras so that the wheels are extremely large.  These features are not including in the small subset dataset of KIT MOMA; thus, this typical wheel loader's features let the model believe it encounters a wheel loader. Based on the analysis, we add some mispredicted samples into the training and validation set, and the AP of wheel loader increases then to 0.6. In this way, we can always rely on a minimal data set to achieve good results on a specific site, though overfitting occurs.

\Figure[ht!](topskip=0pt, botskip=0pt, midskip=0pt)[width=3.3in]{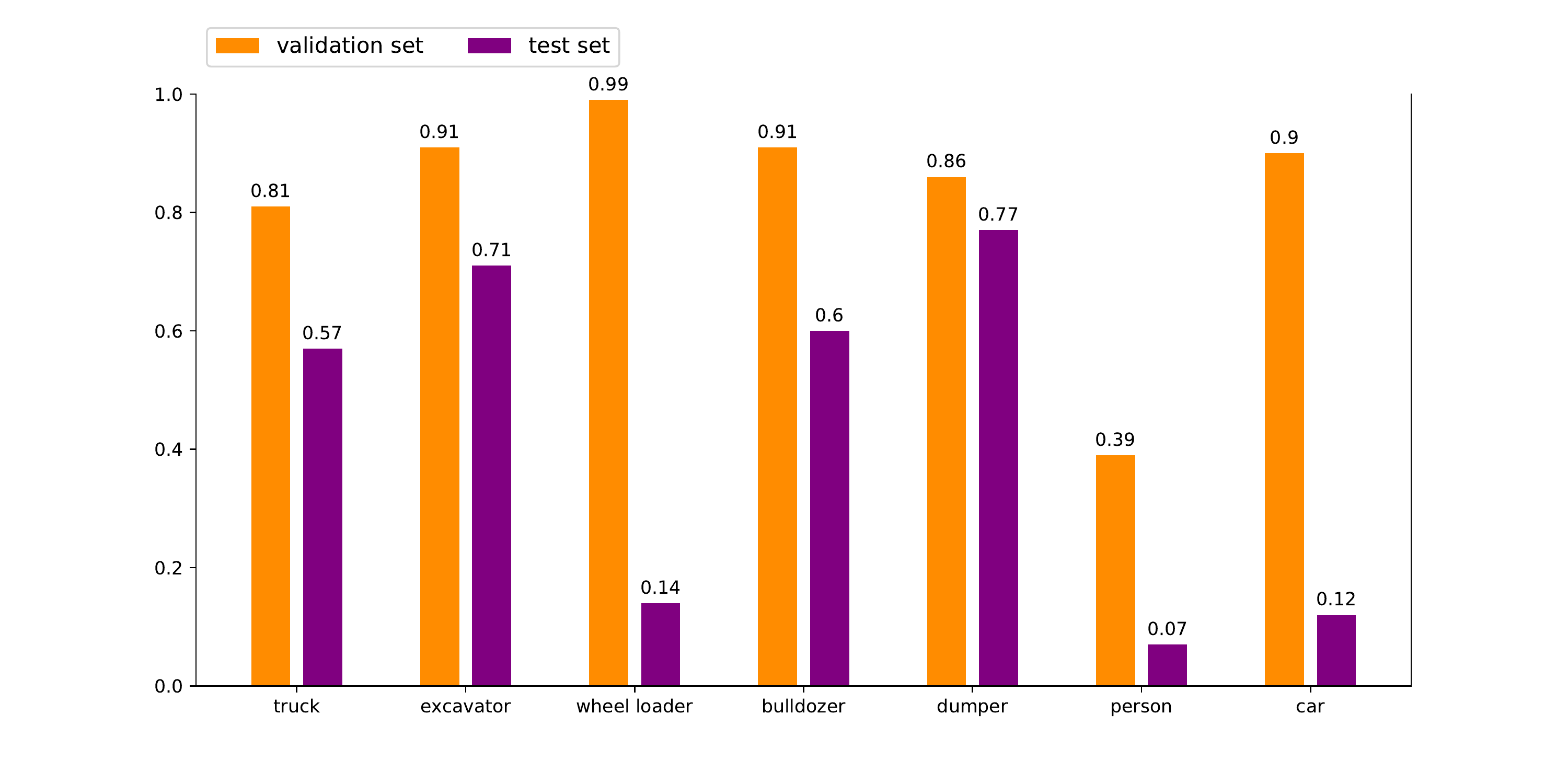}
{Individual prediction AP for each class on other 5,663 images.\label{fig:mAPfor5663}}

\section{Conclusion}

In this work, we presented KIT MOMA, a large-scale and diverse construction machines detection dataset with ground truth label. The dataset is designed to be used as a benchmark for the evaluation of computer vision algorithms to detect mobile machines. Most of the images are gained in real scenarios on the working site, while some other images are downloaded directly from the official website of construction machine companies. Instead of gathering the images in the drivers' view, we collect the samples from the outside view of the mobile machines since we believe it is more in line with the actual situation of autonomous driving of construction machines. 

With our dataset, YOLO v3 is possible to detect mobile machines with mAP of 85\% in general, which is much better than the previous works without using the deep learning algorithms. Notice that we only compared the researchers who have confidently published their code. Also, without considering the instances outside of the specific working site, the mAP goes to almost 90.7\%, which indicates that the predictor is ready for a level four autonomous driving task. Since YOLO is more friendly to real time applications, we recommend adopting this algorithm for the recognition task of construction machines. Finally, recognition performance depends on the dataset quality and how people train the algorithms. By further expanding the data collection and annotation, more satisfying results can be expected. Hence, we also recommend adding the images of interest, such as the excavators or dumpers that are going to be detected, into our KIT MOMA and further train the pretrained model to get a predictor, which is the best suit for the specific level four task.

\subsection{Outlook}

The task of object detection relates to a wide range of knowledge, experience as well as hardware allocations. A further deep study of mobile machines detection algorithms to promote their performance in precision and fps is highly recommended. We hope our work can foster some novel algorithms for detecting mobile machines. Besides algorithmic improvement, some improvements in the dataset can be concluded as follows. In this dataset, the mobile machines are treated as a whole, whereas perceiving component or sub-assembly of mobile machines makes sense in some way as well, for instance, bucket or backhoes of an excavator. In addition, collect extra data of mobile machines in extreme poses if needed. The majority of mobile machines work in normal poses, for instance, an excavator sits on the ground or even in the water, with its bucket moving around; a wheel loader loads coal and unloads it. However, in some situations, machines must work in extreme poses, such as a dumper deposits earth or a wheel loader is buried in the earth but still feebly recognizable. By collecting more images like this may expand the scope of model application. Finally, besides object detection, computer vision is also trending to image segmentation. Pixel-level semantic segmentation can also improve the detection performance of predictors. In our next version, we will publish the dataset with pixel-level annotation of mobile machines.

\bibliography{literature.bib}{}
\bibliographystyle{IEEEtran}

\begin{IEEEbiography}[{\includegraphics[width=1in,height=1.25in,clip,keepaspectratio]{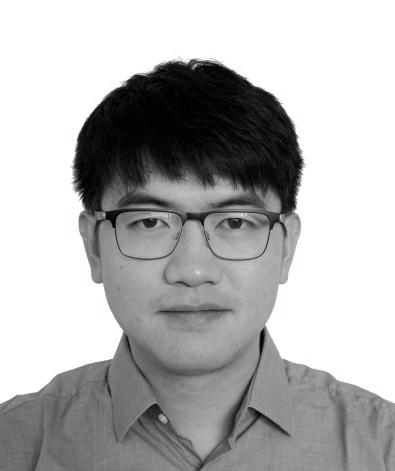}}]{Yusheng Xiang} is pursuing his PhD degree at the Institute of Vehicle System Technology, Karlsruhe Institute of Technology, Karlsruhe, Germany. Also, he is a research scientist at Robert Bosch GmbH, Germany. From Sep. 2020, he is a visiting scholar at the University of California, Berkeley, USA. He received M.Sc. degree in Vehicle Engineering with the focus of Mathematical Model Building and Simulation from the Karlsruhe Institute of Technology, Karlsruhe, in 2017. He has authored 6 influential journal and international conference papers, and holds 4 patents. His group deals with the improvement of mobile machines' performance using Artificial Intelligence and the Internet of Things. 

\end{IEEEbiography}

\begin{IEEEbiography}[{\includegraphics[width=1in,height=1.25in,clip,keepaspectratio]{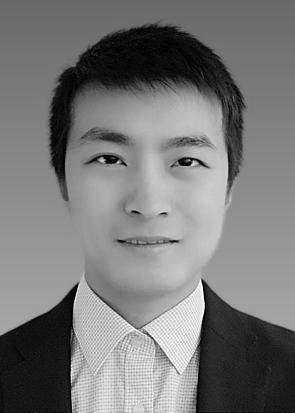}}]{Hongzhe Wang} received B.Sc  from Jilin University, China, and M.Sc in Mechatronics and Vehicle Engineering at Karlsruhe Institute of Technology (KIT), Germany. He has been involved in mobile machine detection with deep-learning frameworks at the Institute of Vehicle System Technology. He just completed the evaluation of Faster-RCNN and YOLOv3 on mobile machines in his master thesis. His research interests include data mining, object detection, machine-learning and deep-learning algorithms.

\end{IEEEbiography}

\begin{IEEEbiography}[{\includegraphics[width=1in,height=1.25in,clip,keepaspectratio]{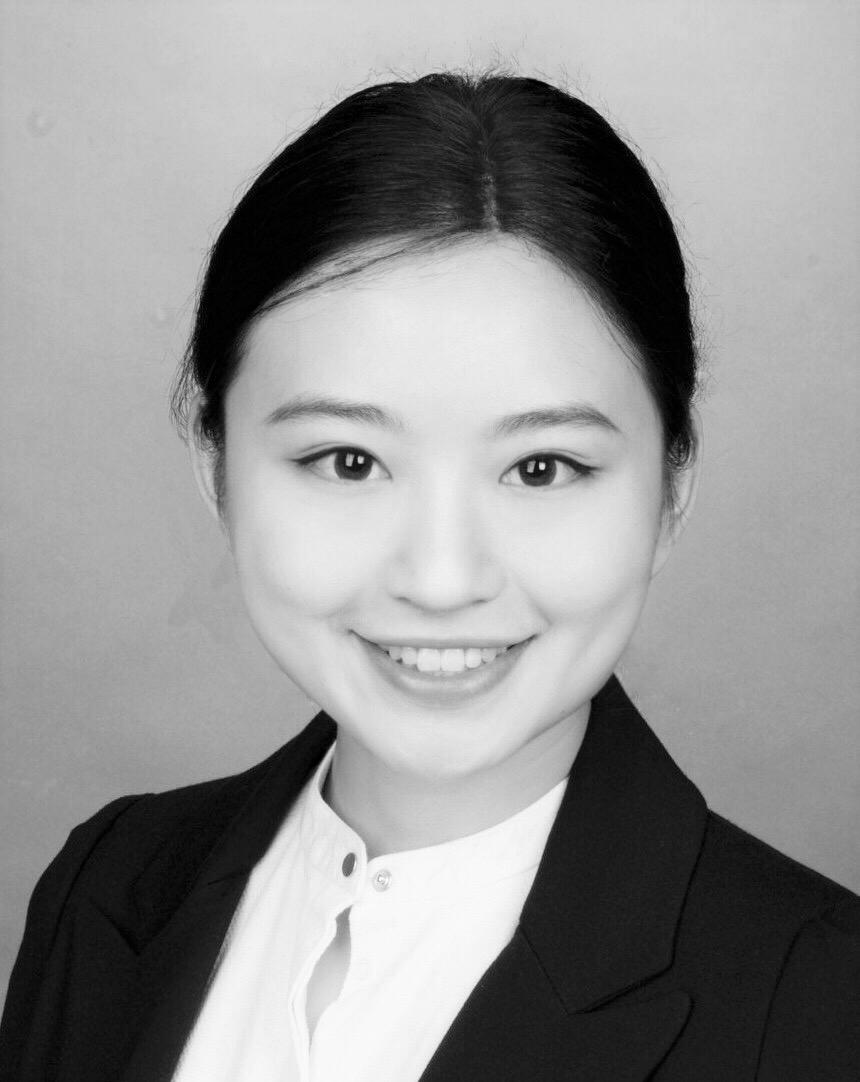}}]{Tianqing Su} received the B.Sc. degree in telecommunication from the Xidian Universtiy, Xi'an, China, in 2013, and the M.Sc. degree in Information Systems Engineering from the Technische Universit\"at Braunschweig, Germany, in 2017. She is currently a Software Engineer in the field of hybrid and electric vehicle at Continental AG, Regensburg, Germany. Before she joined Continental AG, she worked at Robert Bosch GmbH in Abstatt, Germany.

\end{IEEEbiography}

\begin{IEEEbiography}[{\includegraphics[width=1in,height=1.25in,clip,keepaspectratio]{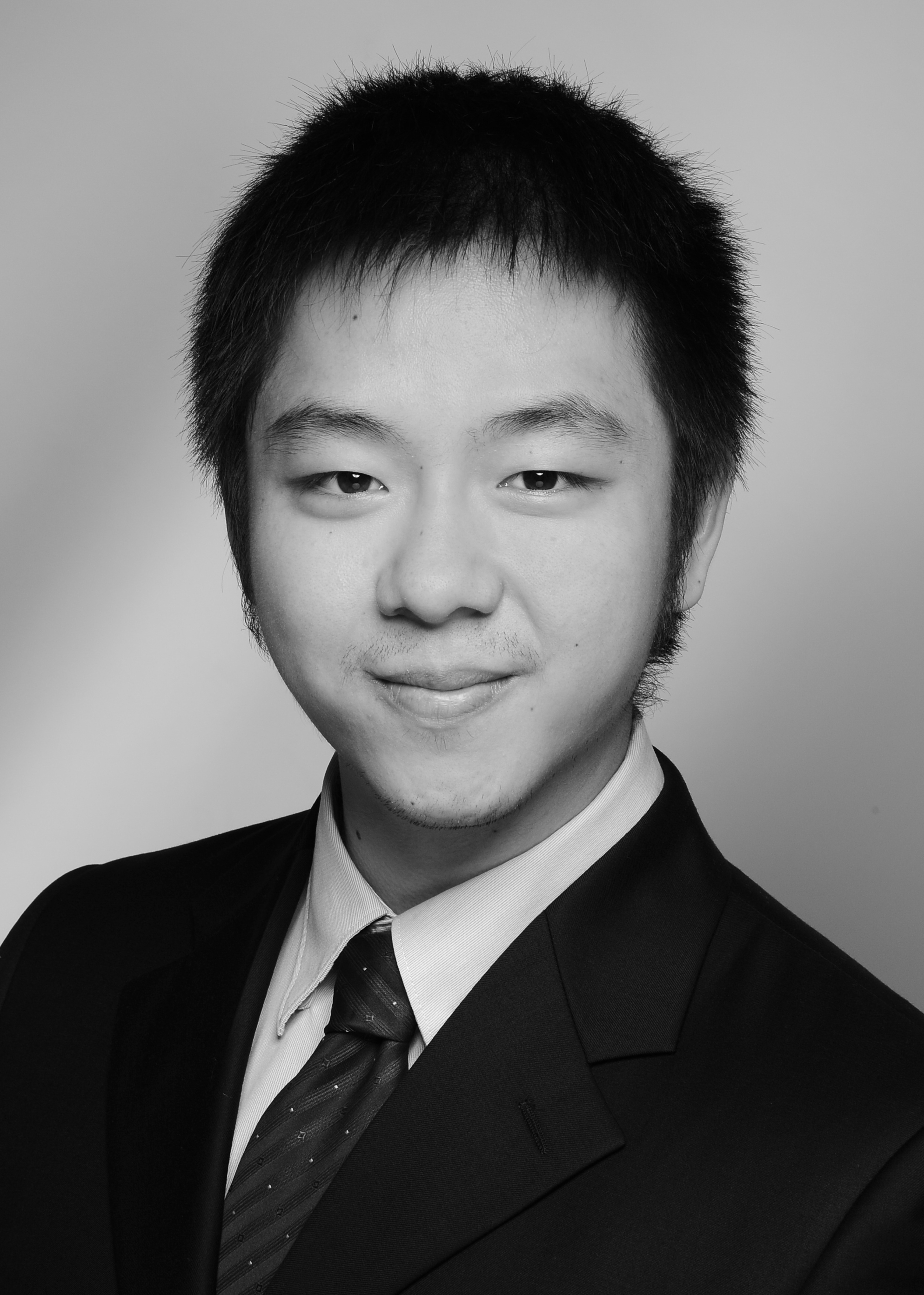}}]{Ruoyu Li} received B.Sc. degree in Mechanical Engineering and Automation from the Beijing Institute of Technology, Peking, China, in 2015, and the M.Sc. degree in Vehicle Engineering and Dynamic {\&} Vibration theory from the Karlsruhe Institute of Technology, Karlsruhe, Germany in 2019. He is committed in modelling and control of dynamic systems and interested in exploring the possibilities for applying Deep Learning and Machine Learning in dynamic systems.

\end{IEEEbiography}

\begin{IEEEbiography}[{\includegraphics[width=1in,height=1.25in,clip,keepaspectratio]{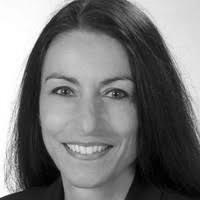}}]{Christine Brach}is the department leader of System Engineering in the division of mobile hydraulics at Robert Bosch GmbH where she leads interdisciplinary teams for large system development projects. As a supervisor for 3 PhD students, she has 15 publications in journals, conferences and patents. . 

\end{IEEEbiography}

\begin{IEEEbiography}[{\includegraphics[width=1in,height=1.25in,clip,keepaspectratio]{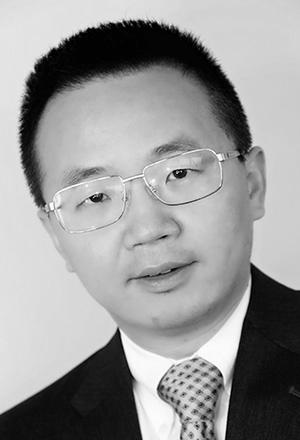}}]{Samuel Mao} received his Ph.D. degree from the University of California at Berkeley in 2000. After that, Prof. Mao started his career at Lawrence Berkeley National Laboratory, where he was a career staff scientist until 2013. He returned to U.C. Berkeley campus as an adjunct professor in 2004, when he also established the Clean Energy Engineering Laboratory that has spun off an international technology development and commercialization institution, the Institute of New Energy, launched in 2013. Having published 180 refereed research articles that have received more than 43,000 citations, Prof. Mao is also an inventor of 80 patents in the U.S. and abroad. He delivered nearly 100 invited, keynote or plenary speeches at international conferences, and previously served as a technical committee member, program review panelist, grant proposal evaluator, and national laboratory observer for the U.S. Department of Energy. In addition to co-founding three international materials and energy technology conferences, he co-chaired Materials Research Society (MRS) annual meeting in the spring of 2011, and the International Conference on Clean Energy in 2012. Prof. Mao received an "R\&D 100" Technology Award (2011) for his technological innovation, and a Berkeley MEGSCO Faculty Teaching Award (2008) for his dedication to higher education.

\end{IEEEbiography}

\begin{IEEEbiography}[{\includegraphics[width=1in,height=1.25in,clip,keepaspectratio]{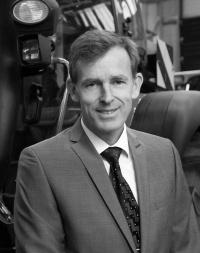}}]{Marcus Geimer} received his Diploma degree in Mechanical Engineering from the RWTH Aachen University, Aachen, Germany in 1990. In 1995, he received his PhD from the Institute of Hydraulics and Pneumatics, today named Institute for Fluid Power Drives and Systems, RWTH Aachen University. He started his industrial career 1995 in the field of construction, where he was the leader of the research group for hydraulic breakers. In 2000, he changed to the hydraulic industry, where he leads the construction and customer development for mobile hydraulics. 
Since 2005, he is a full professor and director at the Institute of Mobile Machines (Mobima), at the Karlsruhe Institute of Technology KIT, Germany. His research activities focus on drives and controls for mobile working machines, like agriculture, construction and municipal vehicles. Research projects on hydrostatic, electric and hybrid drives, as well on traction as on function drives, have been successfully completed. Modern control strategies, like machine learning methods, neural networks or predictive control, are under research. 

\end{IEEEbiography}

\EOD

\end{document}